\documentclass[11pt]{article}

\newcommand{\thetitle}{
Adaptive Signal Resuscitation: Channel-wise Post-Pruning Repair for Sparse Vision Networks
}

\title{\thetitle}

\author{
Qishi Zhan$^{1,* ,\dagger}$ \quad
Ziheng Chen$^{2,* ,\dagger}$ \quad
Minxuan Hu$^{3}$\\[0.6em]
$^{1}$Department of Mathematical and Statistical Sciences, Marquette University, USA\\
$^{2}$The University of Texas at Austin, USA\\
$^{3}$Cornell Ann S. Bowers College of Computing and Information Science, Cornell University, USA\\[0.4em]
\texttt{stokes615@utexas.edu} \quad
\texttt{qishi.zhan@marquette.edu}\\[0.4em]
{\small $^{*}$Equal contribution.}\\
{\small $^{\dagger}$Corresponding authors, in order: Ziheng Chen and Qishi Zhan.}
}

\date{}

\usepackage[margin=1in]{geometry}
\usepackage{times}
\usepackage{microtype}

\usepackage{multirow}
\usepackage{booktabs}
\usepackage{amsmath}
\usepackage{amssymb}
\usepackage{graphicx}
\usepackage[ruled,vlined,linesnumbered]{algorithm2e}
\usepackage{placeins}
\usepackage{float}
\usepackage[colorlinks=true,linkcolor=blue,citecolor=blue,urlcolor=blue]{hyperref}
\usepackage{xcolor}

\begin{document}

\maketitle

\begin{abstract}
One-shot magnitude pruning can cause severe accuracy collapse in the high-sparsity regime, even when the pruning mask preserves the largest weights. We argue that this failure is driven primarily by a granularity mismatch in post-pruning repair, rather than by the reduction in parameter count alone. Under global magnitude pruning, damage is highly heterogeneous across channels within the same layer: nearly collapsed channels can coexist with channels that retain informative activation variance. Existing layer-wise activation repair methods apply a single correction to the whole layer, and can therefore over-amplify damaged channels while trying to restore the layer-level signal. We propose Adaptive Signal Resuscitation (ASR), a training-free channel-wise repair method that matches the granularity of repair to the granularity of damage. ASR estimates a variance-matching correction for each output channel and stabilizes it with a data-driven shrinkage rule, suppressing unreliable corrections for channels with weak post-pruning signal while preserving corrections for healthier channels. Applied before BatchNorm recalibration, ASR requires only forward passes on a small calibration set and no retraining. Across three datasets, four convolutional architectures, and both unstructured and structured sparsity settings, ASR generally improves over layer-wise repair, with the clearest gains in high-sparsity regimes where repair is most needed. On ResNet-50 at 90\% sparsity, ASR recovers 55.6\% top-1 accuracy on CIFAR-10, compared with 41.0\% for layer-wise repair and 28.0\% for BatchNorm-only recalibration. The gains are largest when pruning induces severe and heterogeneous channel damage; in architectures without residual connections, BatchNorm recalibration alone can remain competitive at extreme sparsity. Ablations show that naive channel-wise variance matching is insufficient, and that shrinkage is the key component that stabilizes post-pruning repair.
\end{abstract}

\section{Introduction}
\label{sec:intro}

Unstructured magnitude pruning remains one of the most practical ways
to compress vision networks after training, especially when a trained
dense model must be deployed under strict memory or compute
constraints~\cite{Cheng2024PruningSurvey,Blalock2020StatePruning,Han2015,Han2016DeepCompression,Liu2019RethinkingPruning,Yang2018NetAdapt,Frantar2023}.
A sparse model can be obtained directly from a trained network without
changing the architecture or paying the cost of full retraining. Yet at
extreme sparsity, when 90\% or more of weights are removed, one-shot
magnitude pruning can drive accuracy close to random chance, even when
the surviving weights are those ranked most important by the pruning
rule~\cite{Frankle2019LotteryTicket,Liu2019RethinkingPruning,Renda2020Rewinding}.

This drop in performance is not explained solely by the removal of
important parameters. Global L1 pruning removes weights by magnitude
across the entire network, without respect to layer or channel
boundaries. As a result, the surviving weight density can vary sharply
across channels within the same layer: some channels retain much of
their original mass, while others lose nearly all of it. This
heterogeneous damage causes intermediate activation statistics to shift
non-uniformly across channels and depth, gradually weakening the signal
needed for reliable prediction~\cite{saikumar2025}.

\begin{figure}[t]
    \centering
    \includegraphics[width=\linewidth]{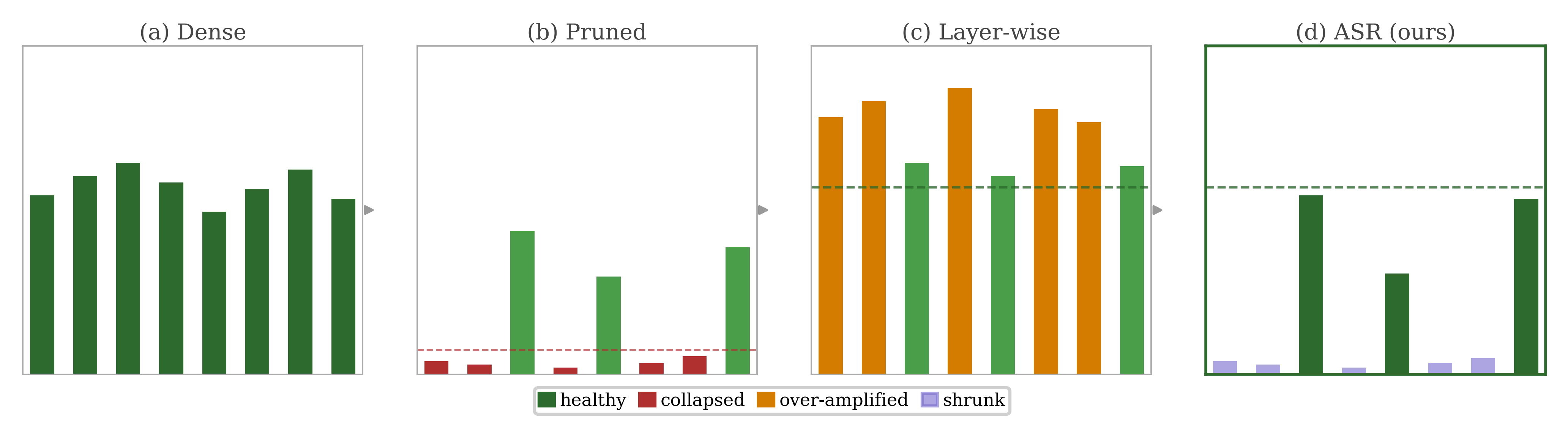}
    \caption{
    A granularity mismatch in post-pruning repair. Global pruning creates
    heterogeneous channel damage within a layer. Layer-wise repair applies
    one shared correction and can over-amplify collapsed channels, whereas
    ASR applies shrinkage-stabilised channel-wise repair.
    }
    \label{fig:hook}
\end{figure}

Repairing this shift without retraining is difficult. A simple baseline
is BatchNorm recalibration, which updates running mean and variance
statistics on a small calibration set after pruning and can recover a
substantial part of the lost accuracy~\cite{Ioffe2015BatchNorm}. A
natural extension is to apply layer-wise activation scaling before
BatchNorm recalibration, matching the variance of each pruned layer to
that of its dense counterpart, as in the recently proposed
REFLOW~\cite{saikumar2025}. This layer-wise strategy, however, assumes
that pruning damage is approximately uniform within each layer. Under
global pruning, that assumption is systematically violated. Severely
damaged and relatively healthy channels can coexist within the same
layer, yet a layer-wise correction applies a shared scalar to all of
them. When collapsed channels contribute disproportionately to the
layer-wise variance ratio, the resulting scalar can become overly large
and amplify damaged channels instead of restoring useful signal.
Layer-wise repair can therefore become less stable than BatchNorm
recalibration alone in the high-sparsity regime where repair is most
needed. Figure~\ref{fig:hook} illustrates this granularity mismatch.

We propose Adaptive Signal Resuscitation (ASR), a training-free
post-pruning repair method that matches the granularity of repair to
the granularity of damage. For each output channel, ASR estimates a
scaling factor from the ratio between dense and pruned activation
variance on a small calibration set, and stabilises this estimate
through a data-driven shrinkage rule. Channels whose variance has
collapsed toward zero receive a correction close to the identity, while
channels with healthier variance retain a stronger correction. This
adaptive suppression prevents the over-amplification of damaged
channels that can destabilise layer-wise scaling. Applied before
BatchNorm recalibration, ASR stabilises the activations entering BN
layers and improves early calibration efficiency, allowing strong
accuracy recovery with fewer recalibration batches.

Our work makes three contributions. We first identify a failure mode of
layer-wise activation repair in high-sparsity pruning: when channel
damage is highly uneven, a shared layer-wise correction can be driven by
collapsed channels and may reduce accuracy below BatchNorm recalibration
alone. We then introduce ASR, a training-free channel-wise repair method
that uses a data-driven shrinkage rule to suppress unreliable corrections
for collapsed channels while retaining stronger corrections for channels
with recoverable signal. Finally, we show across CNN vision benchmarks
that ASR is most beneficial when pruning damage is severe and
heterogeneous, and that applying ASR before BatchNorm recalibration
improves recovery under small calibration budgets. ASR estimates its
repair factors from 64 calibration images using forward evaluation only;
the repaired model is evaluated under the same BatchNorm recalibration
budget as the baselines. The method is designed for convolutional
networks with BatchNorm. Extensions to transformer architectures and
post-training quantization are left for future work. Additional
derivations and diagnostics are provided in Supplementary
Sections~\ref{sec:supp_method}--\ref{sec:supp_nm24_curves}.

\section{Related Work}
\label{sec:related}
A large body of work on neural network pruning studies which weights,
channels, or filters should be removed from a trained model while
preserving predictive performance. Classical saliency-based methods
estimate the importance of weights or connections before removal
~\cite{LeCun1989,Hassibi1993}, while magnitude pruning remains a widely
used practical baseline~\cite{Han2015}. In vision models, both
unstructured and structured pruning have been studied, including filter
pruning, channel pruning, and sparsity-inducing normalisation methods
\cite{Li2017FilterPruning,He2017ChannelPruning,Luo2017ThiNet,Wen2016StructuredSparsity,Liu2017NetworkSlimming,Wang2018StructuredProbabilisticPruning,He2018SoftFilterPruning,Lin2019GAL,Molchanov2019Importance,Liu2019RethinkingPruning}.
More recent methods exploit richer information than weight magnitude
alone~\cite{Frantar2022OBC}. SparseGPT~\cite{Frantar2023} uses approximate second-order
updates to compensate surviving weights, Wanda~\cite{Sun2024} combines
weight magnitude with activation statistics, and LAMP~\cite{Lee2021}
allocates sparsity across layers through a model-level distortion
criterion. These methods primarily decide which weights to remove, how
to allocate sparsity, or how to update the surviving weights. Our work
addresses a different stage: given a fixed pruning mask and fixed
surviving weights, we study how to repair the activation distribution
after pruning without retraining.

Closer to our setting is work on post-pruning signal degradation and
calibration-based repair~\cite{Li2020EagleEye,Lazarevich2021LayerWiseCalibration}.
One-shot pruning can induce progressive variance decay across layers,
leading to signal collapse~\cite{saikumar2025}. A natural post-hoc
repair strategy is therefore to restore layer-wise activation variance
from a small calibration set, without updating trainable weights; the
REFLOW preprint~\cite{saikumar2025} is a recent example of this idea.
Related activation renormalisation methods also appear outside pruning,
for example in REPAIR~\cite{Jordan2023}, which corrects preactivation
statistics when interpolating or merging networks.

These methods share a calibration-based and training-free philosophy,
but their repair granularity is not designed for the heterogeneous
channel collapse caused by global magnitude pruning. In our setting,
healthy and near-dead channels can coexist within the same layer, so a
shared correction can be dominated by collapsed channels and may
over-amplify unreliable signal. ASR keeps the same post-hoc repair
setting, but adapts activation-statistic repair to this pruning-specific
regime by estimating corrections separately for each output channel and
shrinking unreliable corrections toward the identity. This distinction
is important: our ablation in Table~\ref{tab:ablation} shows that raw
channel-wise variance matching is not sufficient. The contribution is
therefore not channel granularity alone, but shrinkage-stabilised
channel-wise repair for collapsed activations.

ASR uses a data-driven shrinkage rule as a post-hoc reliability weight:
after pruning, each channel-wise correction is contracted toward the
identity in proportion to how little post-pruning variance the channel
retains, so that collapsed channels receive minimal adjustment while
healthier channels are corrected more strongly. This differs from the
role of James--Stein shrinkage~\cite{james1961estimation} and Bayesian
shrinkage priors used for regularisation or
compression~\cite{Ishwaran2005,Louizos2017,Molchanov2017}, where
shrinkage is applied during training to induce sparsity in weights or
posteriors. We do not impose a prior on weights or learn a sparse
posterior. In this sense, ASR belongs to the broader family of
small-calibration-set correction methods used in post-training
quantization and pruning~\cite{Nagel2019DataFreeQuantization,Cai2020ZeroQ,Nagel2020,Hubara2021b,Li2021BRECQ,Hubara2021a,Lazarevich2021LayerWiseCalibration},
but focuses on a specific failure mode in high-sparsity convolutional
networks with BatchNorm: layer-wise repair can become unstable when
channel damage is highly heterogeneous.

\section{Adaptive Signal Resuscitation}
\label{sec:method}

\begin{figure}[t]
    \centering
    \includegraphics[width=\linewidth]{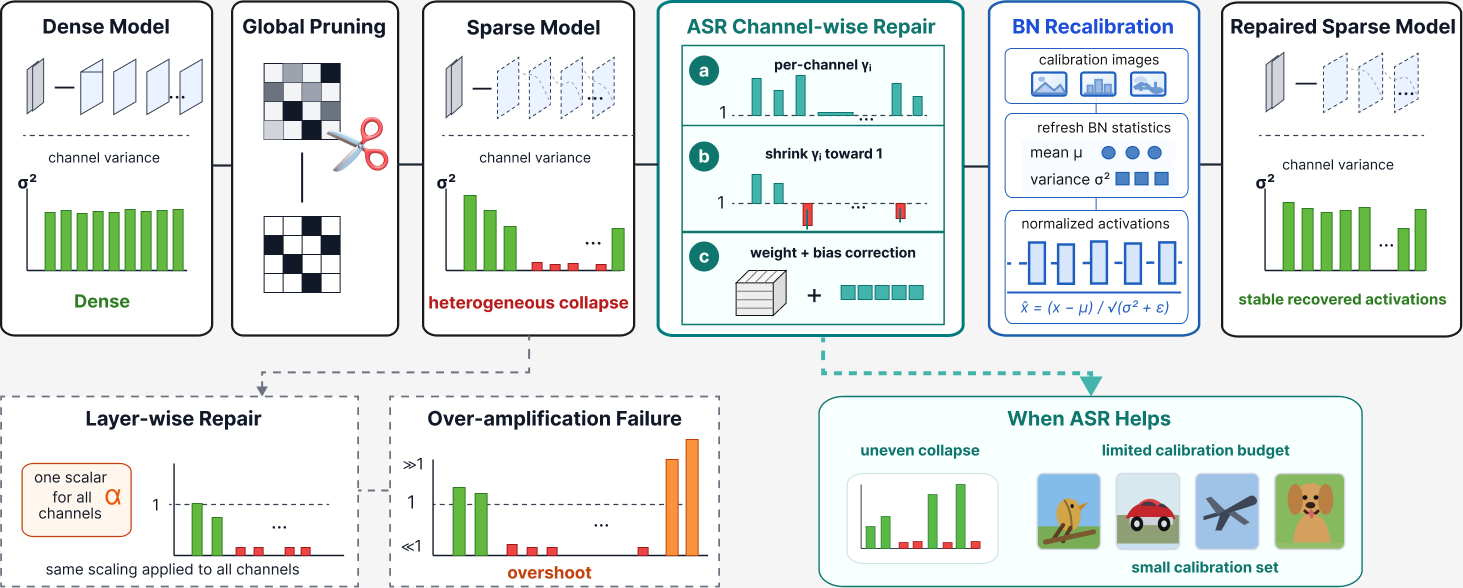}
    \caption{
    Overview of ASR. ASR repairs heterogeneous post-pruning channel collapse
    by applying shrinkage-stabilised channel-wise correction before BatchNorm
    recalibration.
    }
    \label{fig:overview}
\end{figure}

Figure~\ref{fig:overview} summarises the ASR pipeline. Given a dense
model and its pruned counterpart, ASR uses a small calibration set to
measure pre-BatchNorm activation statistics, computes a channel-wise
repair factor for each convolutional output channel, shrinks unreliable
corrections toward the identity, and then recalibrates BatchNorm
statistics under the repaired activation distribution. We first review
why a layer-wise correction can become unstable, then derive the
channel-wise variance-matching estimator, and finally introduce the
shrinkage and bias-correction steps used by ASR.

\subsection{From Layer-wise Repair to Channel-wise Matching}
\label{sec:method_matching}

Let $\mathbf{Y}_d \in \mathbb{R}^{N \times C \times H \times W}$ and
$\mathbf{Y}_p \in \mathbb{R}^{N \times C \times H \times W}$ denote the
pre-BatchNorm output activations of a convolutional layer in the dense
and pruned models, respectively, evaluated on a calibration set of $N$
images. For each output channel $i \in \{1,\dots,C\}$, define
\[
v_{d,i} = \mathrm{Var}(\mathbf{Y}_{d,i}),
\qquad
v_{p,i} = \mathrm{Var}(\mathbf{Y}_{p,i}),
\]
where variance is taken over the batch and spatial dimensions.

A layer-wise repair estimates one correction factor for the whole layer,
\begin{equation}
\gamma_{\mathrm{LW}}
=
\sqrt{
\frac{\frac{1}{C}\sum_{i=1}^{C}\widehat{\mathrm{Var}}(\mathbf{Y}_{d,i})}
     {\frac{1}{C}\sum_{i=1}^{C}\widehat{\mathrm{Var}}(\mathbf{Y}_{p,i})+\epsilon}
}.
\label{eq:gamma_lw}
\end{equation}
This estimator is reasonable when post-pruning damage is relatively
homogeneous within the layer. Under global magnitude pruning, however,
a subset of channels can have near-zero post-pruning variance while
other channels remain active. These near-collapsed channels reduce the
denominator of the layer-wise ratio, but they do not necessarily contain
recoverable signal. The resulting shared correction can therefore become
too large and is applied indiscriminately to both healthy and damaged
channels. This motivates a repair rule whose correction strength is
estimated and stabilised at channel granularity.

If the pruned channel is rescaled by a scalar $\gamma_i$, then its
variance becomes
\[
\mathrm{Var}(\gamma_i \mathbf{Y}_{p,i}) = \gamma_i^2 v_{p,i}.
\]
A natural moment-matching objective is therefore to choose $\gamma_i$ so
that the repaired pruned variance is close to the corresponding dense
variance. We formalise this by the channel-wise criterion
\begin{equation}
\mathcal{L}_i(\gamma_i)
=
\bigl(\gamma_i^2 v_{p,i} - v_{d,i}\bigr)^2.
\label{eq:variance_objective}
\end{equation}
Minimising \eqref{eq:variance_objective} over $\gamma_i \ge 0$ yields
the population target
\begin{equation}
\gamma_i^\star
=
\sqrt{\frac{v_{d,i}}{v_{p,i}}}.
\label{eq:gamma_population}
\end{equation}
The non-negativity constraint $\gamma_i \ge 0$ ensures a unique minimiser:
the objective \eqref{eq:variance_objective} is convex in $\gamma_i^2$,
so the positive square root is the global minimum over the feasible set,
provided $v_{d,i} > 0$.

Replacing population variances by sample variances from the calibration
set gives the raw variance-matching estimator
\begin{equation}
\hat{\gamma}_i
=
\sqrt{
\frac{\widehat{\mathrm{Var}}(\mathbf{Y}_{d,i})}
{\widehat{\mathrm{Var}}(\mathbf{Y}_{p,i}) + \epsilon}
},
\label{eq:gamma_raw}
\end{equation}
where $\epsilon > 0$ is a numerical floor.

Equation~\eqref{eq:gamma_raw} is well motivated when
$\widehat{\mathrm{Var}}(\mathbf{Y}_{p,i})$ is stably estimated. Under
high sparsity, however, some channels may become severely damaged after
global pruning, and their post-pruning variances can approach zero. In
that regime, the denominator in \eqref{eq:gamma_raw} becomes small,
making the raw correction factor unstable. Thus, channel granularity
alone is not sufficient: the raw estimator is least reliable for the
channels that are most severely damaged, and applying it directly may
amplify residual noise instead of restoring useful signal.

\subsection{Empirical Bayes Shrinkage Toward the Identity}
\label{sec:method_shrinkage}

To stabilise the channel-wise corrections, we introduce a data-driven
shrinkage step that contracts unreliable estimates
toward the identity map $\gamma_i = 1$. The shrinkage baseline is chosen
from the empirical distribution of post-pruning channel variances within
the layer:
\begin{equation}
\lambda
=
\mathrm{median}\!\left(
\left\{
\widehat{\mathrm{Var}}(\mathbf{Y}_{p,i})
\right\}_{i=1}^{C}
\right).
\label{eq:prior}
\end{equation}
We use the median as the default layer-level shrinkage baseline because
it is less sensitive to anomalous channel variances than the mean and
performs well in the high-sparsity regime studied in our ablation.

We then define the shrinkage weight
\begin{equation}
s_i
=
\frac{\widehat{\mathrm{Var}}(\mathbf{Y}_{p,i})}
     {\widehat{\mathrm{Var}}(\mathbf{Y}_{p,i}) + \lambda},
\qquad 0 \le s_i < 1,
\label{eq:shrinkage_weight}
\end{equation}
and the final repair factor
\begin{equation}
\gamma_i
=
s_i \hat{\gamma}_i + (1-s_i)\cdot 1.
\label{eq:gamma_final}
\end{equation}
Equivalently,
\begin{equation}
\gamma_i - 1
=
s_i(\hat{\gamma}_i - 1),
\label{eq:gamma_interp}
\end{equation}
so $\gamma_i$ always lies between the raw estimate $\hat{\gamma}_i$ and
the identity correction $1$. The identity correction is a conservative
target: ASR does not attempt to revive a channel whose post-pruning
activation has nearly vanished, since such a channel is more likely to
carry amplified residual noise than useful signal.

The behaviour of \eqref{eq:gamma_final} follows directly from
\eqref{eq:shrinkage_weight}. If
$\widehat{\mathrm{Var}}(\mathbf{Y}_{p,i}) \to 0$, then $s_i \to 0$ and
\[
\gamma_i \to 1,
\]
so the method applies little or no multiplicative correction to a nearly
collapsed channel. If
$\widehat{\mathrm{Var}}(\mathbf{Y}_{p,i}) \gg \lambda$, then $s_i \to 1$
and
\[
\gamma_i \to \hat{\gamma}_i,
\]
so the estimator approaches the full variance-matching correction. In
this sense, $s_i$ acts as a continuous reliability weight derived from
the observed post-pruning activation scale of each channel.

For a convolutional kernel
$\mathbf{W}_\ell \in \mathbb{R}^{C_{\mathrm{out}} \times C_{\mathrm{in}} \times k \times k}$,
the repair is applied per output channel:
\begin{equation}
\widetilde{\mathbf{W}}_{\ell,i,:,:,:}
=
\gamma_i \mathbf{W}_{\ell,i,:,:,:},
\qquad i=1,\dots,C_{\mathrm{out}}.
\label{eq:weight_rescale}
\end{equation}
Thus each output channel is repaired by a scale factor determined by its
own post-pruning activation statistics rather than by a layer-wide
average.

\subsection{Bias Correction and BatchNorm Recalibration}
\label{sec:method_bn}

Pruning can shift activation means as well as variances~\cite{Nagel2019DataFreeQuantization}.
After channel-wise rescaling, we optionally adjust the bias so that the
repaired pruned activations match the dense activations at the first
moment. Let
$\boldsymbol{\gamma} = (\gamma_1,\dots,\gamma_C)^\top$ and denote by
$\widehat{\mathbb{E}}[\mathbf{Y}_d]$ and
$\widehat{\mathbb{E}}[\mathbf{Y}_p]$ the per-channel sample means over
the batch and spatial dimensions. If the repaired pre-BN activations are
written as
\[
\widetilde{\mathbf{Y}}_p
=
\boldsymbol{\gamma} \odot \mathbf{Y}_p
+
(\mathbf{b}_{\mathrm{new}} - \mathbf{b}_{\mathrm{old}}),
\]
then imposing the moment-matching condition
\[
\widehat{\mathbb{E}}[\widetilde{\mathbf{Y}}_p]
=
\widehat{\mathbb{E}}[\mathbf{Y}_d]
\]
gives
\begin{equation}
\mathbf{b}_{\mathrm{new}}
=
\mathbf{b}_{\mathrm{old}}
+
\widehat{\mathbb{E}}[\mathbf{Y}_d]
-
\boldsymbol{\gamma} \odot \widehat{\mathbb{E}}[\mathbf{Y}_p].
\label{eq:bias}
\end{equation}
This correction aligns the first moment after multiplicative repair,
addressing the mean shift that pruning introduces independently of the
variance distortion.

For networks with BatchNorm layers, we apply ASR before BatchNorm
recalibration. The channel-wise rescaling and bias adjustment first
repair the pre-BN activation distribution, after which the BatchNorm
running statistics are recomputed on the calibration set. This ordering
separates two roles: ASR repairs the convolutional signal at channel
granularity, while BatchNorm recalibration updates the downstream
normalisation statistics under the repaired activation distribution.

The full ASR procedure requires only forward passes on the calibration
set to collect activation statistics, followed by deterministic weight
and bias updates. It uses no gradient computation and does not retrain
the sparse model. Supplementary Section~\ref{sec:supp_method} gives
the corresponding activation-shift derivation and a formal stability
view of the shrinkage rule.

\begin{algorithm}[!t]
\SetAlgoLined
\SetKwInOut{Input}{Input}
\SetKwInOut{Output}{Output}
\SetKwInOut{Hyper}{Hyperparam}
\DontPrintSemicolon
\SetKwComment{Comment}{$\triangleright$\ }{}

\caption{Adaptive Signal Resuscitation (ASR)}
\label{alg:asr}

\Input{Dense model $f_d$; pruned model $f_p$; calibration set $\mathcal{X} = \{x_n\}_{n=1}^{N}$}
\Hyper{Numerical floor $\epsilon > 0$}
\Output{Repaired pruned model $f_p$}
\BlankLine

\tcp{Stage 1: Channel-wise data-driven shrinkage repair}
\ForEach{convolutional layer $\ell = 2, \ldots, L$}{
    \tcp*[l]{layer 1 skipped: its input is raw pixels, whose statistics
    are independent of pruning and require no repair}
    collect pre-BN activations
    $\mathbf{Y}_d \leftarrow f_d^{(\ell)}(\mathcal{X})$ and
    $\mathbf{Y}_p \leftarrow f_p^{(\ell)}(\mathcal{X})$
    \Comment*[r]{$\in \mathbb{R}^{N \times C \times H \times W}$}
    \BlankLine

    \For{$i = 1$ \KwTo $C$}{
        $v_{d,i} \leftarrow \widehat{\mathrm{Var}}(\mathbf{Y}_{d,i})$
        \Comment*[r]{dense channel variance}

        $v_{p,i} \leftarrow \widehat{\mathrm{Var}}(\mathbf{Y}_{p,i})$
        \Comment*[r]{pruned channel variance}

        $\hat{\gamma}_i \leftarrow \sqrt{v_{d,i} / (v_{p,i} + \epsilon)}$
        \Comment*[r]{raw variance-matching correction}
    }
    \BlankLine

    $\lambda \leftarrow \mathrm{median}\!\left(\{v_{p,i}\}_{i=1}^{C}\right)$
    \Comment*[r]{data-driven shrinkage baseline}
    \BlankLine

    \For{$i = 1$ \KwTo $C$}{
        $s_i \leftarrow v_{p,i} / (v_{p,i} + \lambda)$
        \Comment*[r]{shrinkage weight}

        $\gamma_i \leftarrow s_i \hat{\gamma}_i + (1 - s_i)$
        \Comment*[r]{shrunk correction}
    }
    \BlankLine

    rescale output-channel filters:
    $\mathbf{W}_{\ell,i,:,:,:} \leftarrow \gamma_i \mathbf{W}_{\ell,i,:,:,:}$
    for $i=1,\dots,C$
    \Comment*[r]{channel-wise weight repair}
    \BlankLine

    $\boldsymbol{\mu}_d \leftarrow \widehat{\mathbb{E}}[\mathbf{Y}_d]$
    and
    $\boldsymbol{\mu}_p \leftarrow \widehat{\mathbb{E}}[\mathbf{Y}_p]$
    \Comment*[r]{per-channel means}

    $\mathbf{b}_\ell \leftarrow
    \mathbf{b}_\ell + \boldsymbol{\mu}_d - \boldsymbol{\gamma} \odot \boldsymbol{\mu}_p$
    \Comment*[r]{bias correction}
}
\BlankLine

\tcp{Stage 2: BatchNorm recalibration}
Re-estimate BatchNorm running statistics of $f_p$ on $\mathcal{X}$\;
\Return $f_p$
\end{algorithm}

\section{Experiments}
\label{sec:experiments}

We design the experiments to evaluate whether ASR addresses the failure
mode identified above: unstable layer-wise repair under heterogeneous
channel damage. In particular, we ask whether ASR improves over
layer-wise repair in high-sparsity regimes, whether the effect persists
across architectures and datasets, and whether applying ASR before
BatchNorm recalibration improves accuracy recovery under limited
calibration budgets.

We evaluate ASR across three datasets, four architectures, four pruning
conditions, and three repair methods. The datasets are CIFAR-10,
CIFAR-100~\cite{krizhevsky2009learning}, and
Imagenette~\cite{Howard2019Imagenette}, covering different levels of
task difficulty and class granularity. The architectures are
ResNet-18, ResNet-50~\cite{he2016deep},
DenseNet-121~\cite{huang2017densely}, and
VGG-16-BN~\cite{simonyan2015very}, which span different depths,
connectivity patterns, and BatchNorm densities. All models are
initialised from ImageNet-pretrained weights~\cite{Deng2009ImageNet}
and fine-tuned for five epochs on the target dataset, with all layers
frozen except the final residual block and classification head. This
protocol creates a controlled post-training pruning setting while
keeping most of the pretrained visual representation fixed.

We consider global unstructured L1 magnitude pruning at 50\%, 70\%, and
90\% sparsity, implemented with
\texttt{torch.nn.utils.prune.global\_unstructured}, as well as
structured 2:4 (N:M) sparsity~\cite{Mishra2021AcceleratingSparseDNNs,Zhou2021NM,Hubara2021a}.
After pruning, masks are made permanent before any repair is applied.
We compare three post-pruning repair strategies. BN Only resets and
re-estimates BatchNorm running mean and variance statistics on a
calibration set. LW+BN applies layer-wise activation scaling in the
style of REFLOW~\cite{saikumar2025} before the same BatchNorm
recalibration. ASR+BN applies our channel-wise shrinkage-stabilised
repair before BatchNorm recalibration. We also report the
unpruned dense model and the unrepaired pruned model as upper and lower
reference points.

ASR estimates its channel-wise correction factors from a small
calibration subset of 64 training images. After this repair step, we
study BatchNorm recalibration efficiency by varying the number of
recalibration batches over $\{10,20,30,50\}$ with batch size 128, and
report top-1 test accuracy as a function of calibration budget. All
calibration images are drawn from the training set and are disjoint from
the test set. Unless otherwise stated, main results are reported at
recalibration step $b=20$, which provides an early but stable operating
point for comparing repair methods. Experiments are run on a single
NVIDIA T4 GPU.

\paragraph{Implementation details.}
All repair methods are applied after the pruning mask has been fixed;
none changes the sparsity pattern or performs gradient-based
fine-tuning. ASR estimates its channel-wise repair factors using 64
training images. BatchNorm recalibration uses the calibration budget
specified in each experiment, and all repair methods are evaluated under
the same recalibration budget. Calibration images are sampled from the
training set and are disjoint from the test set. Results use a fixed
pruning and calibration seed for controlled comparison across repair
methods. Supplementary Section~\ref{sec:step20} provides the empirical
rationale for reporting the main comparison at $b{=}20$.

\section{Results}
\label{sec:results}

\subsection{Main Results}
\label{sec:main_results}

\begin{figure}[H]
    \centering
    \includegraphics[width=\linewidth]{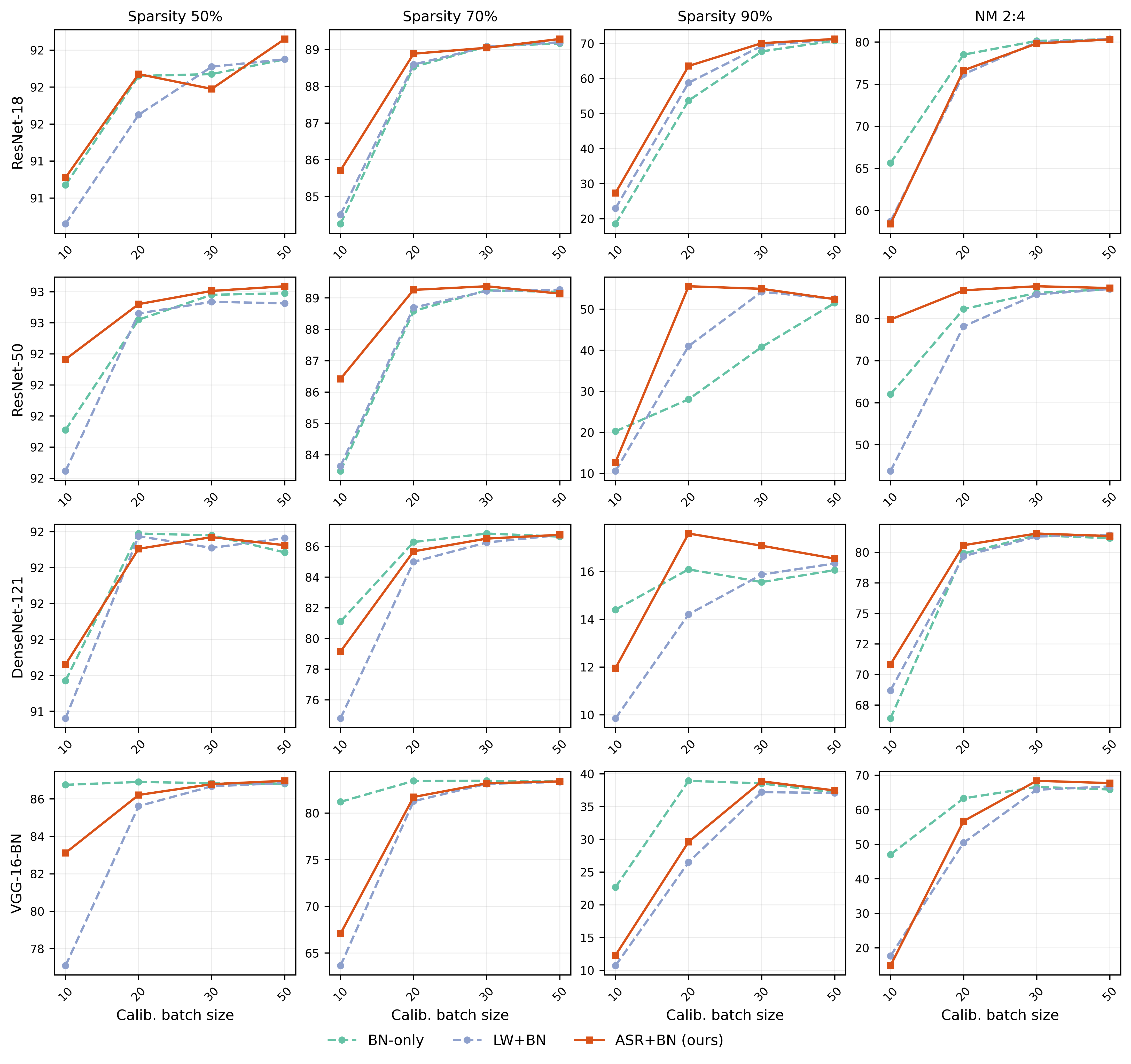}
    \caption{
    Top-1 accuracy versus calibration batch size on CIFAR-10. ASR+BN
    generally leads BN Only and LW+BN at small calibration budgets, with
    the largest gains appearing in higher-sparsity settings.
    }
    \label{fig:cifar10}
\end{figure}

Table~\ref{tab:main} reports top-1 test accuracy at calibration step
$b{=}20$ across all architecture, dataset, and pruning combinations.
Figure~\ref{fig:cifar10} shows the corresponding accuracy trajectories
across calibration batch sizes on CIFAR-10. A recurring pattern across
settings is the instability of layer-wise scaling at high sparsity. In
multiple configurations, LW+BN underperforms the simpler BN-only
baseline, which is consistent with the view that a single per-layer
scalar can be overly influenced by severely damaged channels and
therefore over-correct the repaired activations. The effect is most
pronounced on VGG-16-BN at 90\% sparsity, where LW+BN falls more than
12 percentage points below BN-only on CIFAR-10 and nearly 9 points on
CIFAR-100. A similar pattern appears on DenseNet-121, where LW+BN
underperforms BN-only at 90\% sparsity across all three datasets.
DenseNet's dense connectivity causes each layer's output to be
concatenated with all preceding feature maps, so weight density---and
therefore channel variance---varies more sharply across layers under
global pruning than in residual networks; this may compound the variance
distortion that layer-wise repair struggles to correct.

\begin{table}[!t]
\centering
\caption{
Top-1 accuracy (\%) at calibration step $b{=}20$, batch size 128. Bold
indicates the best result among repair methods at each setting. The
unrepaired pruned model is shown in gray for reference. Results are
shown for sparsity levels 70\%, 90\%, and NM~2:4; 50\% results are
omitted because all repair methods perform similarly at low sparsity.
}
\label{tab:main}
\setlength{\tabcolsep}{3.6pt}
\scriptsize
\begin{tabular}{llccc ccc ccc}
\toprule
& & \multicolumn{3}{c}{CIFAR-10}
  & \multicolumn{3}{c}{CIFAR-100}
  & \multicolumn{3}{c}{Imagenette} \\
\cmidrule(lr){3-5}\cmidrule(lr){6-8}\cmidrule(lr){9-11}
Architecture & Method & 70\% & 90\% & 2:4
                     & 70\% & 90\% & 2:4
                     & 70\% & 90\% & 2:4 \\
\midrule
\multirow{5}{*}{ResNet-18}
  & Dense      & \multicolumn{3}{c}{\textcolor{gray}{92.77}} & \multicolumn{3}{c}{\textcolor{gray}{75.56}} & \multicolumn{3}{c}{\textcolor{gray}{97.27}} \\
  & No Repair  & 37.73 & 8.19 & 15.68 & 3.20 & 1.00 & 3.83 & 92.60 & 11.12 & 32.33 \\
  & BN Only    & 88.53 & 53.70 & \textbf{78.50} & 60.45 & \textbf{15.55} & \textbf{48.75} & 96.69 & 90.04 & 91.59 \\
  & LW+BN      & 88.59 & 58.77 & 76.17 & 60.36 & 7.10 & 44.46 & 96.69 & 91.06 & 92.22 \\
  & ASR+BN     & \textbf{88.88} & \textbf{63.56} & 76.64 & \textbf{60.75} & 15.25 & 46.39 & \textbf{96.73} & \textbf{91.68} & \textbf{92.47} \\
\midrule
\multirow{5}{*}{ResNet-50}
  & Dense      & \multicolumn{3}{c}{\textcolor{gray}{93.33}} & \multicolumn{3}{c}{\textcolor{gray}{75.05}} & \multicolumn{3}{c}{\textcolor{gray}{98.22}} \\
  & No Repair  & 10.87 & 10.00 & 13.55 & 1.17 & 1.00 & 2.31 & 93.22 & 22.05 & 37.18 \\
  & BN Only    & 88.58 & 28.04 & 82.28 & 58.18 & 4.01 & 57.22 & 97.52 & \textbf{86.39} & 93.92 \\
  & LW+BN      & 88.69 & 41.01 & 78.18 & 58.02 & 5.04 & 51.06 & \textbf{97.68} & 66.15 & 95.31 \\
  & ASR+BN     & \textbf{89.25} & \textbf{55.59} & \textbf{86.74} & \textbf{58.80} & \textbf{6.88} & \textbf{60.75} & 97.68 & 82.98 & \textbf{95.67} \\
\midrule
\multirow{5}{*}{DenseNet-121}
  & Dense      & \multicolumn{3}{c}{\textcolor{gray}{92.75}} & \multicolumn{3}{c}{\textcolor{gray}{75.11}} & \multicolumn{3}{c}{\textcolor{gray}{98.88}} \\
  & No Repair  & 40.74 & 9.29 & 12.57 & 8.86 & 0.93 & 2.16 & 78.90 & 14.79 & 49.97 \\
  & BN Only    & \textbf{86.28} & 16.09 & 79.91 & \textbf{62.09} & 1.96 & 53.17 & \textbf{97.34} & \textbf{73.10} & 95.87 \\
  & LW+BN      & 85.00 & 14.21 & 79.68 & 58.84 & 2.25 & 53.31 & 96.75 & 29.54 & 95.44 \\
  & ASR+BN     & 85.68 & \textbf{17.59} & \textbf{80.58} & 60.65 & \textbf{2.99} & \textbf{54.00} & 97.10 & 44.13 & \textbf{96.25} \\
\midrule
\multirow{5}{*}{VGG-16-BN}
  & Dense      & \multicolumn{3}{c}{\textcolor{gray}{87.35}} & \multicolumn{3}{c}{\textcolor{gray}{64.61}} & \multicolumn{3}{c}{\textcolor{gray}{98.50}} \\
  & No Repair  & 52.15 & 11.90 & 14.72 & 24.26 & 1.34 & 5.03 & 92.20 & 15.86 & 62.99 \\
  & BN Only    & \textbf{83.44} & \textbf{38.93} & \textbf{63.29} & \textbf{57.72} & \textbf{14.74} & \textbf{35.50} & \textbf{97.41} & \textbf{70.29} & 94.10 \\
  & LW+BN      & 81.27 & 26.49 & 50.43 & 53.02 & 6.06 & 21.59 & 96.89 & 47.43 & \textbf{94.25} \\
  & ASR+BN     & 81.72 & 29.62 & 56.70 & 53.27 & 5.89 & 27.15 & 97.06 & 52.04 & 94.05 \\
\bottomrule
\end{tabular}
\end{table}

ASR+BN usually improves over LW+BN, with the largest gains appearing at
the highest sparsity levels. The largest absolute gain over LW+BN occurs
on ResNet-50 at 90\% sparsity on CIFAR-10, where ASR+BN reaches
55.59\% compared to 41.01\% for LW+BN, an improvement of 14.58
percentage points. At lower sparsities, where channel collapse is milder
and more homogeneous, the differences between repair methods are
correspondingly smaller. This pattern suggests that the main benefit of
ASR is not stronger rescaling alone, but selective rescaling in settings
where channel damage is highly uneven.

The comparison between ASR+BN and BN-only is more nuanced. On ResNet-18
and ResNet-50, ASR+BN matches or exceeds BN-only in most high-sparsity
settings. On VGG-16-BN, however, BN-only outperforms both LW+BN and
ASR+BN at 90\% sparsity and under NM~2:4 sparsity. VGG-16-BN lacks
residual connections, so activation signal cannot bypass collapsed
layers; when a large fraction of channels approach zero variance, even
shrinkage-stabilised corrections operate on a distribution that carries
little recoverable information. In this regime, multiplicative repair
at any granularity appears to introduce more instability than it
resolves, and BatchNorm recalibration alone---which makes no
multiplicative adjustment to the convolutional weights---proves the
more conservative and reliable strategy. Whether channel-wise repair
can be further stabilised in very deep networks without skip connections
remains an open question. On Imagenette, where the fine-tuned dense
model is already close to ceiling performance, all three repair methods
perform more similarly, consistent with a setting in which the
post-pruning damage is less severe.

Figure~\ref{fig:cifar10} also shows that the advantage of ASR+BN over
LW+BN is most visible at small calibration budgets, especially at
$b{=}10$ and $b{=}20$. In this regime, repairing convolutional weights
before BatchNorm recalibration appears to stabilise the activations
entering BN layers and thereby improve early calibration efficiency. As
$b$ increases, the gap between methods narrows and performance tends to
approach a common ceiling, consistent with BatchNorm statistics becoming
better estimated across all repair strategies.

\subsection{Mechanism: Channel-wise Variance Repair}
\label{sec:mechanism}

\begin{figure}[H]
    \centering
    \includegraphics[width=\linewidth]{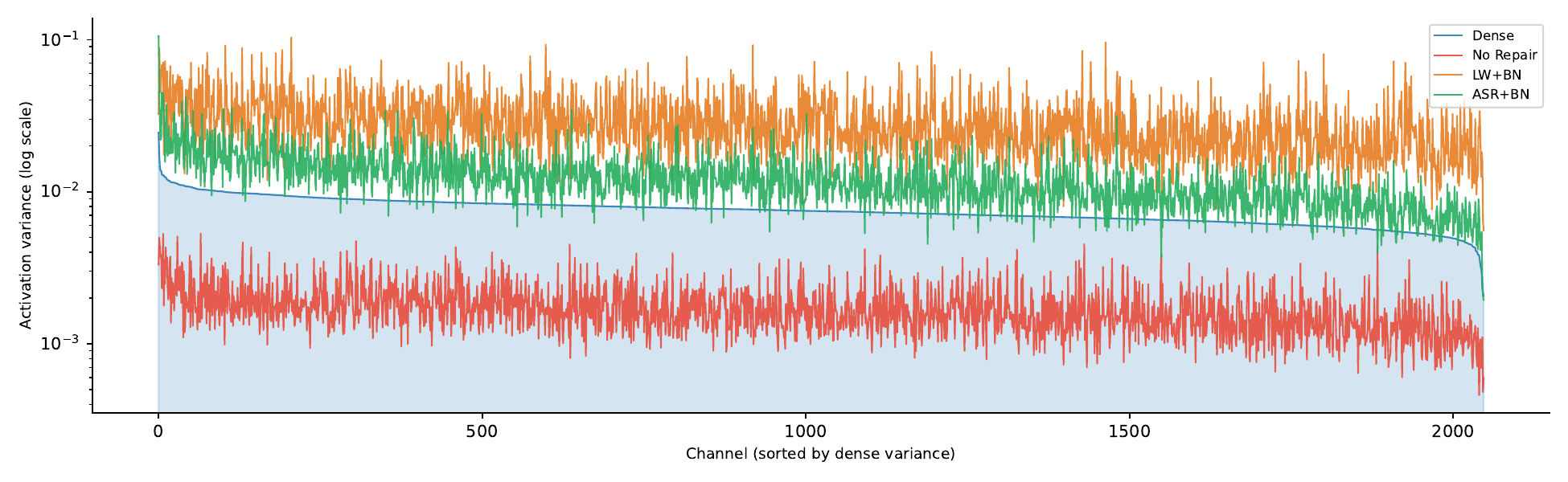}
    \caption{
    Per-channel activation variance after repair for ResNet-50 on CIFAR-10
    with NM~2:4 sparsity at \texttt{layer4.0.conv3}. The dense curve is the
    reference. No Repair remains under-corrected ($\alpha=0.23$), LW+BN
    over-corrects ($\alpha=3.84$), and ASR+BN is closer to the dense target
    ($\alpha=1.68$).
    }
\label{fig:signal_collapse}
\end{figure}

Figure~\ref{fig:signal_collapse} visualises per-channel activation
variance after each repair method for ResNet-50 on CIFAR-10 with NM~2:4
sparsity. The channels are sorted by dense variance, and each repaired
variance is compared against the dense reference. The ideal repair
corresponds to a log--log slope $\alpha=1$, meaning that repaired
variance tracks the dense reference proportionally across all channels;
a slope above one indicates systematic over-correction, and a slope
below one indicates under-correction.

The pruned model remains strongly under-corrected ($\alpha=0.23$), with
most channels lying well below the dense level. Layer-wise scaling moves
in the opposite direction and substantially over-corrects the activations
($\alpha=3.84$), pushing many channel variances above the dense
reference; a slope well above one means that channels with low
post-pruning variance receive the same large multiplicative correction
as healthy channels, amplifying noise rather than signal. ASR+BN
produces a slope closer to the ideal value ($\alpha=1.68$), indicating
a more moderate correction that is better aligned with the dense target.
This tighter variance alignment translates directly into the accuracy
differences observed in Table~\ref{tab:main}: on ResNet-50 at NM~2:4
sparsity, ASR+BN reaches 86.74\% on CIFAR-10 compared with 82.28\% for
BN-only and 78.18\% for LW+BN, a gap that is consistent with the
over-correction visible in Figure~\ref{fig:signal_collapse}. More
broadly, this pattern confirms the argument in
Section~\ref{sec:method}: a single layer-wise scalar is too coarse to
reflect the heterogeneous damage induced by global pruning, whereas
channel-wise correction with shrinkage yields a more stable repair.

\begin{figure}[H]
    \centering
    \includegraphics[width=\linewidth]{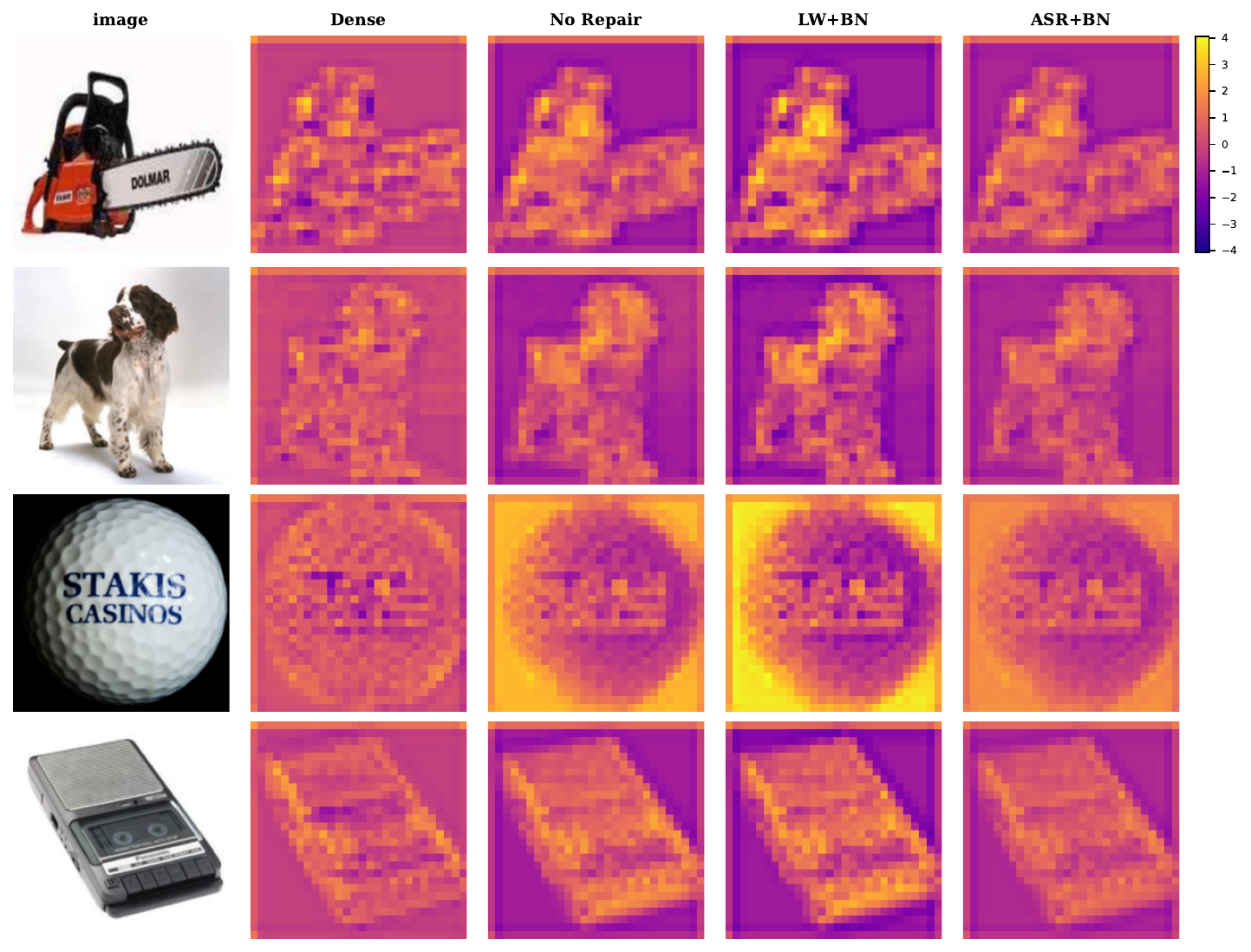}
    \caption{
    Spatial activation maps at \texttt{layer2.0.conv1}, channel~1 of
    ResNet-18 on Imagenette under 90\% global L1 sparsity. Each row
    corresponds to a different test image. Pruning weakens the feature
    response, LW+BN over-amplifies it relative to the dense reference,
    and ASR+BN recovers a spatial pattern closer to the dense
    activation.
    }
    \label{fig:activation_maps}
\end{figure}

Figure~\ref{fig:activation_maps} gives a spatial view of the same
failure mode. After pruning, the feature response is weakened; after
layer-wise repair, the response can be over-amplified relative to the
dense reference. ASR+BN produces activation maps that are visually
closer to the dense model, consistent with its more selective
channel-wise correction. Supplementary
Section~\ref{sec:supp_layer_stats} provides layer-wise heatmaps that
show the same over-correction pattern across representative networks.

\subsection{When Does ASR Help?}
\label{sec:when_asr_helps}

\begin{figure}[H]
    \centering
    \includegraphics[width=\linewidth]{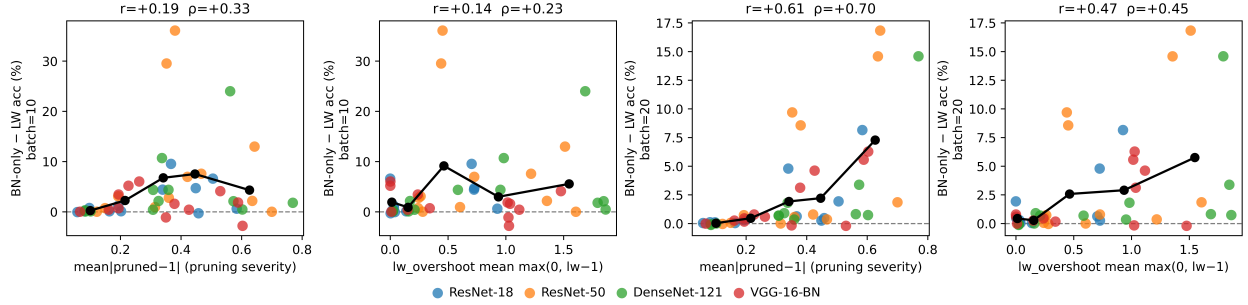}
    \caption{
    Accuracy gain of ASR+BN over LW+BN versus pruning severity
    (average channel variance deviation from the dense reference) at
    calibration step $b{=}20$. Each point is one architecture--dataset--sparsity
    combination. Larger severity is associated with larger ASR advantage
    ($r=0.61$, $\rho=0.70$).
    }
    \label{fig:asr_advantage}
\end{figure}

The results above suggest that ASR is most useful when pruning induces
severe channel-wise variance distortion. We quantify this relationship
by comparing the accuracy gain of ASR+BN over LW+BN at calibration step
$b{=}20$ against a pruning-severity statistic computed from channel-wise
variance ratios. As shown in Figure~\ref{fig:asr_advantage}, larger
deviation of pruned channel variances from the dense reference is
associated with a larger ASR advantage over layer-wise repair, with
Pearson correlation $r=0.61$ and Spearman correlation $\rho=0.70$.

This trend is consistent with the mechanism proposed in
Section~\ref{sec:method}. When channel damage is mild, a shared
layer-wise correction is often sufficient and the gain from ASR is
small. As post-pruning channel variances become more distorted, however,
the layer-wise scalar becomes increasingly mismatched to the
heterogeneous channel state. In this regime, ASR benefits from
estimating correction strength at channel granularity and shrinking
unreliable corrections toward the identity. A small number of points in
Figure~\ref{fig:asr_advantage} fall near or marginally below zero,
corresponding to low-severity settings where the two methods perform
comparably; these are concentrated at pruning severity below~0.2, where
channel collapse is mild enough that a shared scalar already captures
most of the needed correction.
\subsection{Ablation Study}
\label{sec:ablation}

Table~\ref{tab:ablation} isolates the contributions of channel
granularity, data-driven shrinkage, and bias correction under a
representative high-sparsity setting. We compare BN only, layer-wise
repair, raw channel-wise variance matching without shrinkage, and
several ASR variants. This ablation separates three effects: whether
channel granularity alone is sufficient, whether shrinkage stabilises
raw channel-wise correction, and whether bias correction provides an
additional gain once shrinkage is used.

The results show that channel granularity alone is not sufficient in
this setting. Moving from layer-wise repair to raw channel-wise repair
does not improve accuracy, indicating that naive channel-wise correction
remains unstable at high sparsity. Adding shrinkage leads to a large
gain: the fixed-prior ASR variant already improves over both layer-wise
and raw channel-wise repair, while the adaptive variants perform
substantially better. Among them, the best result is achieved by ASR
with a median prior and bias correction. The mean-prior variant is also
strong, and the gap between the median-prior variants with and without
bias correction is relatively small, suggesting that the choice of prior
has a secondary effect once the shrinkage mechanism is in place. Taken
together, these results confirm that the main benefit of ASR comes from
stabilised channel-wise repair, while bias correction plays a supporting
role.

\begin{table}[H]
\centering
\caption{
Ablation on ResNet-50/CIFAR-10 at 90\% sparsity. Top-1 accuracy (\%)
is reported for layer-wise repair, raw channel-wise repair, and ASR
variants.
}
\label{tab:ablation}
\small
\begin{tabular}{lc}
\toprule
Method & Accuracy \\
\midrule
BN only                              & 29.11 \\
Layer-wise + BN                      & 39.42 \\
Channel-wise raw + BN                & 29.30 \\
ASR, fixed prior                     & 45.24 \\
ASR, median prior                    & 52.26 \\
ASR, mean prior + bias correction    & 53.95 \\
ASR, median prior + bias correction  & \textbf{55.59} \\
\bottomrule
\end{tabular}
\end{table}

This ablation clarifies why ASR is not simply direct activation
renormalisation at a finer granularity. Moving from layer-wise to raw
channel-wise matching is insufficient because collapsed channels make
unshrunk variance ratios unreliable. The gain appears when channel-wise
repair is made reliability-aware through shrinkage. ASR is therefore a
collapse-aware repair rule for fixed sparse models, rather than a
stronger form of unconstrained activation matching.
\section{Conclusion}
\label{sec:conclusion}

We have presented Adaptive Signal Resuscitation (ASR), a training-free
post-pruning repair method for activation distribution shift in sparse
vision networks. The central observation is that post-pruning damage is
not uniformly distributed within a layer. Under global magnitude
pruning, severely collapsed channels can coexist with channels that
retain informative activation variance, making a single layer-wise
correction too coarse. In this regime, layer-wise scaling can be
dominated by collapsed channels and may over-amplify unreliable signal
rather than restore useful activations. ASR addresses this granularity
mismatch by estimating channel-wise variance-matching corrections and
stabilising them with a data-driven shrinkage rule, which pulls
unreliable corrections toward the identity while preserving stronger
adjustments for channels with sufficient remaining signal.

Across four convolutional architectures, three datasets, and both
unstructured and structured sparsity settings, ASR+BN generally
improves over layer-wise repair, with the clearest gains in the
high-sparsity regimes where repair is most needed. The gains are
largest when pruning induces severe channel-wise variance distortion,
supporting the view that ASR is most useful when layer-wise repair
becomes mismatched to heterogeneous channel damage. We also find that
applying ASR before BatchNorm recalibration improves early calibration
efficiency, allowing strong accuracy recovery under small calibration
budgets without gradient updates or retraining.

\section{Limitations and Future Work}
\label{sec:limitations}

The scope of this study points to several natural extensions. ASR
requires access to the dense model's activations during calibration,
which matches many post-training compression workflows but may not hold
when only a pruned model is retained. Our fine-tuning protocol keeps
most pretrained layers fixed, providing a controlled setting for
isolating post-pruning repair effects rather than an exhaustive
evaluation of all fully fine-tuned deployment pipelines.

The method is designed for convolutional networks with BatchNorm.
Extending the same granularity-matched repair principle to architectures
with different normalisation mechanisms, such as transformer models with
layer normalisation, is a promising direction. The results on
VGG-16-BN further suggest that multiplicative channel-wise repair loses
effectiveness when skip connections are absent and channel collapse is
especially severe, pointing to an informative boundary case for future
repair criteria.

\FloatBarrier
\clearpage
\bibliographystyle{plain}
\bibliography{ref}

@inproceedings{Han2015,
  author = {Han, Song and Pool, Jeff and Tran, John and Dally, William J.},
  title = {Learning both Weights and Connections for Efficient Neural Networks},
  booktitle = {Advances in Neural Information Processing Systems},
  volume = {28},
  publisher = {Curran Associates, Inc.},
  year = {2015}
}

@inproceedings{LeCun1989,
  author = {LeCun, Yann and Denker, John S. and Solla, Sara A.},
  title = {Optimal Brain Damage},
  booktitle = {Advances in Neural Information Processing Systems},
  volume = {2},
  publisher = {Morgan-Kaufmann},
  year = {1989}
}

@inproceedings{Hassibi1993,
  author = {Hassibi, B. and Stork, D. G. and Wolff, G. J.},
  title = {Optimal Brain Surgeon and General Network Pruning},
  booktitle = {IEEE International Conference on Neural Networks},
  pages = {293--299},
  year = {1993},
  doi = {10.1109/ICNN.1993.298572}
}

@inproceedings{Frantar2023,
  author = {Frantar, Elias and Alistarh, Dan},
  title = {SparseGPT: Massive Language Models Can Be Accurately Pruned in One-Shot},
  booktitle = {Proceedings of the 40th International Conference on Machine Learning},
  series = {ICML},
  articleno = {414},
  numpages = {15},
  year = {2023}
}

@inproceedings{Sun2024,
  author = {Sun, Mingjie and Liu, Zhuang and Bair, Anna and Kolter, J. Zico},
  title = {A Simple and Effective Pruning Approach for Large Language Models},
  booktitle = {The Twelfth International Conference on Learning Representations},
  year = {2024}
}

@inproceedings{Lee2021,
  author = {Lee, Jaeho and Park, Sejun and Mo, Sangwoo and Ahn, Sungsoo and Shin, Jinwoo},
  title = {Layer-Adaptive Sparsity for the Magnitude-Based Pruning},
  booktitle = {International Conference on Learning Representations},
  year = {2021}
}

@inproceedings{Jordan2023,
  author = {Jordan, Keller and Sedghi, Hanie and Saukh, Olga and Entezari, Rahim and Neyshabur, Behnam},
  title = {{REPAIR}: {RE}normalizing Permuted Activations for Interpolation Repair},
  booktitle = {The Eleventh International Conference on Learning Representations},
  year = {2023}
}

@inproceedings{Nagel2020,
  author = {Nagel, Markus and Amjad, Rana Ali and Van Baalen, Mart and Louizos, Christos and Blankevoort, Tijmen},
  title = {Up or Down? Adaptive Rounding for Post-Training Quantization},
  booktitle = {Proceedings of the 37th International Conference on Machine Learning},
  series = {Proceedings of Machine Learning Research},
  volume = {119},
  pages = {7197--7206},
  publisher = {PMLR},
  year = {2020}
}

@inproceedings{Hubara2021a,
  author = {Hubara, Itay and Chmiel, Brian and Island, Moshe and Banner, Ron and Naor, Joseph and Soudry, Daniel},
  title = {Accelerated Sparse Neural Training: A Provable and Efficient Method to Find {N:M} Transposable Masks},
  booktitle = {Advances in Neural Information Processing Systems},
  volume = {34},
  pages = {21099--21111},
  publisher = {Curran Associates, Inc.},
  year = {2021}
}

@inproceedings{Hubara2021b,
  author = {Hubara, Itay and Nahshan, Yury and Hanani, Yair and Banner, Ron and Soudry, Daniel},
  title = {Accurate Post Training Quantization With Small Calibration Sets},
  booktitle = {Proceedings of the 38th International Conference on Machine Learning},
  series = {Proceedings of Machine Learning Research},
  volume = {139},
  pages = {4466--4475},
  publisher = {PMLR},
  year = {2021}
}

@article{james1961estimation,
  author = {James, William and Stein, Charles},
  title = {Estimation with Quadratic Loss},
  journal = {Proceedings of the Fourth Berkeley Symposium on Mathematical Statistics and Probability},
  volume = {1},
  pages = {361--379},
  year = {1961}
}

@article{Ishwaran2005,
  author = {Ishwaran, Hemant and Rao, J. Sunil},
  title = {Spike and Slab Variable Selection: Frequentist and Bayesian Strategies},
  journal = {The Annals of Statistics},
  volume = {33},
  number = {2},
  pages = {730--773},
  publisher = {Institute of Mathematical Statistics},
  year = {2005},
  doi = {10.1214/009053604000001147}
}

@inproceedings{he2016deep,
  author = {He, Kaiming and Zhang, Xiangyu and Ren, Shaoqing and Sun, Jian},
  title = {Deep Residual Learning for Image Recognition},
  booktitle = {2016 IEEE Conference on Computer Vision and Pattern Recognition},
  pages = {770--778},
  publisher = {IEEE Computer Society},
  year = {2016},
  doi = {10.1109/CVPR.2016.90}
}

@inproceedings{huang2017densely,
  author = {Huang, Gao and Liu, Zhuang and Van Der Maaten, Laurens and Weinberger, Kilian Q.},
  title = {Densely Connected Convolutional Networks},
  booktitle = {Proceedings of the IEEE Conference on Computer Vision and Pattern Recognition},
  pages = {4700--4708},
  year = {2017}
}

@inproceedings{simonyan2015very,
  author = {Simonyan, Karen and Zisserman, Andrew},
  title = {Very Deep Convolutional Networks for Large-Scale Image Recognition},
  booktitle = {International Conference on Learning Representations},
  year = {2015},
  url = {https://arxiv.org/abs/1409.1556}
}

@techreport{krizhevsky2009learning,
  author = {Krizhevsky, Alex},
  title = {Learning Multiple Layers of Features from Tiny Images},
  institution = {University of Toronto},
  year = {2009},
  url = {https://www.cs.toronto.edu/~kriz/learning-features-2009-TR.pdf}
}

@misc{Howard2019Imagenette,
  author = {Howard, Jeremy},
  title = {Imagenette: A Smaller Subset of 10 Easily Classified Classes from ImageNet},
  year = {2019},
  howpublished = {\url{https://github.com/fastai/imagenette}},
  note = {Accessed: 2026-05-05}
}

@inproceedings{Deng2009ImageNet,
  author = {Deng, Jia and Dong, Wei and Socher, Richard and Li, Li-Jia and Li, Kai and Li, Fei-Fei},
  title = {ImageNet: A Large-Scale Hierarchical Image Database},
  booktitle = {2009 IEEE Computer Society Conference on Computer Vision and Pattern Recognition Workshops},
  pages = {248--255},
  publisher = {IEEE Computer Society},
  year = {2009},
  doi = {10.1109/CVPR.2009.5206848}
}

@article{Cheng2024PruningSurvey,
  author = {Cheng, Hongrong and Zhang, Miao and Shi, Javen Qinfeng},
  title = {A Survey on Deep Neural Network Pruning: Taxonomy, Comparison, Analysis, and Recommendations},
  journal = {IEEE Transactions on Pattern Analysis and Machine Intelligence},
  volume = {46},
  number = {12},
  pages = {10558--10578},
  year = {2024},
  doi = {10.1109/TPAMI.2024.3447085}
}

@inproceedings{Ioffe2015BatchNorm,
  author = {Ioffe, Sergey and Szegedy, Christian},
  title = {Batch Normalization: Accelerating Deep Network Training by Reducing Internal Covariate Shift},
  booktitle = {Proceedings of the 32nd International Conference on Machine Learning},
  series = {Proceedings of Machine Learning Research},
  volume = {37},
  pages = {448--456},
  publisher = {PMLR},
  year = {2015}
}

@inproceedings{Li2017FilterPruning,
  author = {Li, Hao and Kadav, Asim and Durdanovic, Igor and Samet, Hanan and Graf, Hans Peter},
  title = {Pruning Filters for Efficient ConvNets},
  booktitle = {International Conference on Learning Representations},
  year = {2017}
}

@inproceedings{He2017ChannelPruning,
  author = {He, Yihui and Zhang, Xiangyu and Sun, Jian},
  title = {Channel Pruning for Accelerating Very Deep Neural Networks},
  booktitle = {2017 IEEE International Conference on Computer Vision},
  pages = {1398--1406},
  year = {2017},
  doi = {10.1109/ICCV.2017.155}
}

@inproceedings{Luo2017ThiNet,
  author = {Luo, Jian-Hao and Wu, Jianxin and Lin, Weiyao},
  title = {ThiNet: A Filter Level Pruning Method for Deep Neural Network Compression},
  booktitle = {2017 IEEE International Conference on Computer Vision},
  pages = {5068--5076},
  year = {2017},
  doi = {10.1109/ICCV.2017.541}
}

@inproceedings{Liu2017NetworkSlimming,
  author = {Liu, Zhuang and Li, Jianguo and Shen, Zhiqiang and Huang, Gao and Yan, Shoumeng and Zhang, Changshui},
  title = {Learning Efficient Convolutional Networks through Network Slimming},
  booktitle = {2017 IEEE International Conference on Computer Vision},
  pages = {2755--2763},
  year = {2017},
  doi = {10.1109/ICCV.2017.298}
}

@inproceedings{Molchanov2019Importance,
  author = {Molchanov, Pavlo and Mallya, Arun and Tyree, Stephen and Frosio, Iuri and Kautz, Jan},
  title = {Importance Estimation for Neural Network Pruning},
  booktitle = {2019 IEEE/CVF Conference on Computer Vision and Pattern Recognition},
  pages = {11256--11264},
  publisher = {IEEE Computer Society},
  year = {2019},
  doi = {10.1109/CVPR.2019.01152}
}

@inproceedings{Lazarevich2021LayerWiseCalibration,
  author = {Lazarevich, Ivan and Kozlov, Alexander and Malinin, Nikita},
  title = {Post-Training Deep Neural Network Pruning via Layer-Wise Calibration},
  booktitle = {Proceedings of the IEEE/CVF International Conference on Computer Vision Workshops},
  pages = {798--805},
  year = {2021}
}

@inproceedings{Louizos2017,
  author = {Louizos, Christos and Ullrich, Karen and Welling, Max},
  title = {Bayesian Compression for Deep Learning},
  booktitle = {Advances in Neural Information Processing Systems},
  volume = {30},
  publisher = {Curran Associates, Inc.},
  year = {2017}
}

@inproceedings{Molchanov2017,
  author = {Molchanov, Dmitry and Ashukha, Arsenii and Vetrov, Dmitry},
  title = {Variational Dropout Sparsifies Deep Neural Networks},
  booktitle = {Proceedings of the 34th International Conference on Machine Learning},
  series = {Proceedings of Machine Learning Research},
  volume = {70},
  pages = {2498--2507},
  publisher = {PMLR},
  year = {2017}
}

@misc{saikumar2025,
  author = {Saikumar, Dhananjay and Varghese, Blesson},
  title = {Signal Collapse in One-Shot Pruning: When Sparse Models Fail to Distinguish Neural Representations},
  year = {2025},
  eprint = {2502.15790},
  archivePrefix = {arXiv},
  primaryClass = {cs.LG},
  url = {https://arxiv.org/abs/2502.15790}
}

@inproceedings{Liu2019RethinkingPruning,
title={Rethinking the Value of Network Pruning},
author={Zhuang Liu and Mingjie Sun and Tinghui Zhou and Gao Huang and Trevor Darrell},
booktitle={International Conference on Learning Representations},
year={2019},
url={https://openreview.net/forum?id=rJlnB3C5Ym},
}

@inproceedings{Frantar2022OBC,
 author = {Frantar, Elias and Alistarh, Dan},
 booktitle = {Advances in Neural Information Processing Systems},
 pages = {4475--4488},
 publisher = {Curran Associates, Inc.},
 title = {Optimal Brain Compression: A Framework for Accurate Post-Training Quantization and Pruning},
 url = {https://proceedings.neurips.cc/paper_files/paper/2022/file/1caf09c9f4e6b0150b06a07e77f2710c-Paper-Conference.pdf},
 volume = {35},
 year = {2022}
}

@inproceedings{Nagel2019DataFreeQuantization,
  author={Nagel, Markus and Baalen, Mart Van and Blankevoort, Tijmen and Welling, Max},
  booktitle={2019 IEEE/CVF International Conference on Computer Vision (ICCV)}, 
  title={Data-Free Quantization Through Weight Equalization and Bias Correction}, 
  year={2019},
  volume={},
  number={},
  pages={1325-1334},
  doi={10.1109/ICCV.2019.00141}
}

@inproceedings{Li2021BRECQ,
title={{\{}BRECQ{\}}: Pushing the Limit of Post-Training Quantization by Block Reconstruction},
author={Yuhang Li and Ruihao Gong and Xu Tan and Yang Yang and Peng Hu and Qi Zhang and Fengwei Yu and Wei Wang and Shi Gu},
booktitle={International Conference on Learning Representations},
year={2021},
url={https://openreview.net/forum?id=POWv6hDd9XH}
}

@misc{Mishra2021AcceleratingSparseDNNs,
      title={Accelerating Sparse Deep Neural Networks}, 
      author={Asit Mishra and Jorge Albericio Latorre and Jeff Pool and Darko Stosic and Dusan Stosic and Ganesh Venkatesh and Chong Yu and Paulius Micikevicius},
      year={2021},
      eprint={2104.08378},
      archivePrefix={arXiv},
      primaryClass={cs.LG},
      url={https://arxiv.org/abs/2104.08378}, 
}

@inproceedings{Frankle2019LotteryTicket,
title={The Lottery Ticket Hypothesis: Finding Sparse, Trainable Neural Networks},
author={Jonathan Frankle and Michael Carbin},
booktitle={International Conference on Learning Representations},
year={2019},
url={https://openreview.net/forum?id=rJl-b3RcF7},
}

@article{Blalock2020StatePruning,
  author = {Blalock, Davis and Gonzalez Ortiz, Jose Javier and Frankle, Jonathan and Guttag, John},
  title = {What is the State of Neural Network Pruning?},
  journal = {Proceedings of Machine Learning and Systems},
  volume = {2},
  pages = {129--146},
  year = {2020}
}

@inproceedings{Han2016DeepCompression,
  author = {Han, Song and Mao, Huizi and Dally, William J.},
  title = {Deep Compression: Compressing Deep Neural Networks with Pruning, Trained Quantization and Huffman Coding},
  booktitle = {International Conference on Learning Representations},
  year = {2016},
  url = {https://arxiv.org/abs/1510.00149}
}

@inproceedings{Renda2020Rewinding,
  author = {Renda, Alex and Frankle, Jonathan and Carbin, Michael},
  title = {Comparing Rewinding and Fine-Tuning in Neural Network Pruning},
  booktitle = {International Conference on Learning Representations},
  year = {2020},
  url = {https://openreview.net/forum?id=S1gSj0NKvB}
}

@inproceedings{Wen2016StructuredSparsity,
  author = {Wen, Wei and Wu, Chunpeng and Wang, Yandan and Chen, Yiran and Li, Hai},
  title = {Learning Structured Sparsity in Deep Neural Networks},
  booktitle = {Advances in Neural Information Processing Systems},
  volume = {29},
  year = {2016},
  url = {https://proceedings.neurips.cc/paper_files/paper/2016/hash/41bfd20a38bb1b0bec75acf0845530a7-Abstract.html}
}

@inproceedings{Wang2018StructuredProbabilisticPruning,
  author = {Wang, Huan and Zhang, Qiming and Wang, Yuehai and Hu, Haoji},
  title = {Structured Probabilistic Pruning for Convolutional Neural Network Acceleration},
  booktitle = {British Machine Vision Conference},
  year = {2018},
  url = {https://bmvc2018.org/contents/papers/0870.pdf}
}

@inproceedings{He2018SoftFilterPruning,
  author = {He, Yang and Kang, Guoliang and Dong, Xuanyi and Fu, Yanwei and Yang, Yi},
  title = {Soft Filter Pruning for Accelerating Deep Convolutional Neural Networks},
  booktitle = {Proceedings of the Twenty-Seventh International Joint Conference on Artificial Intelligence},
  pages = {2234--2240},
  year = {2018},
  doi = {10.24963/ijcai.2018/309}
}

@inproceedings{Lin2019GAL,
  author = {Lin, Shaohui and Ji, Rongrong and Yan, Chenqian and Zhang, Baochang and Cao, Liujuan and Ye, Qixiang and Huang, Feiyue and Doermann, David},
  title = {Towards Optimal Structured CNN Pruning via Generative Adversarial Learning},
  booktitle = {Proceedings of the IEEE/CVF Conference on Computer Vision and Pattern Recognition},
  month = {June},
  year = {2019},
  url = {https://openaccess.thecvf.com/content_CVPR_2019/html/Lin_Towards_Optimal_Structured_CNN_Pruning_via_Generative_Adversarial_Learning_CVPR_2019_paper.html}
}

@inproceedings{Li2020EagleEye,
  author = {Li, Bailin and Wu, Bowen and Su, Jiang and Wang, Guangrun},
  title = {EagleEye: Fast Sub-Net Evaluation for Efficient Neural Network Pruning},
  booktitle = {Computer Vision -- ECCV 2020},
  series = {Lecture Notes in Computer Science},
  volume = {12347},
  pages = {639--654},
  publisher = {Springer},
  year = {2020},
  doi = {10.1007/978-3-030-58536-5_38}
}

@inproceedings{Cai2020ZeroQ,
  author = {Cai, Yaohui and Yao, Zhewei and Dong, Zhen and Gholami, Amir and Mahoney, Michael W. and Keutzer, Kurt},
  title = {ZeroQ: A Novel Zero Shot Quantization Framework},
  booktitle = {Proceedings of the IEEE/CVF Conference on Computer Vision and Pattern Recognition},
  pages = {13169--13178},
  year = {2020},
  url = {https://openaccess.thecvf.com/content_CVPR_2020/papers/Cai_ZeroQ_A_Novel_Zero_Shot_Quantization_Framework_CVPR_2020_paper.pdf}
}

@inproceedings{Zhou2021NM,
  author = {Zhou, Aojun and Ma, Yukun and Zhu, Junnan and Liu, Jianbo and Zhang, Zhijie and Yuan, Kun and Sun, Wenxiu and Li, Hongsheng},
  title = {Learning N:M Fine-Grained Structured Sparse Neural Networks from Scratch},
  booktitle = {International Conference on Learning Representations},
  year = {2021},
  url = {https://arxiv.org/abs/2102.04010}
}

@inproceedings{Yang2018NetAdapt,
  author = {Yang, Tien-Ju and Howard, Andrew and Chen, Bo and Zhang, Xiao and Go, Alec and Sandler, Mark and Sze, Vivienne and Adam, Hartwig},
  title = {NetAdapt: Platform-Aware Neural Network Adaptation for Mobile Applications},
  booktitle = {Proceedings of the European Conference on Computer Vision},
  month = {September},
  year = {2018},
  url = {https://openaccess.thecvf.com/content_ECCV_2018/html/Tien-Ju_Yang_NetAdapt_Platform-Aware_Neural_ECCV_2018_paper.html}
}

\clearpage
\appendix
\clearpage

\section{Methodology}
\label{sec:supp_method}

This section expands the method introduced in Section~\ref{sec:method} of the main paper.
We provide additional methodological details for Adaptive Signal
Resuscitation (ASR). The main paper presents ASR as a training-free
post-pruning repair method that corrects activation distribution shift
at channel granularity. This supplement expands the derivation and gives
a formal account of why a shared layer-wise correction can become
unstable when pruning damage is heterogeneous across channels.

\subsection{Activation Distribution Shift from Pruning}

Consider a convolutional layer with weight tensor
$\mathbf{W} \in \mathbb{R}^{C_{\mathrm{out}} \times C_{\mathrm{in}} \times k \times k}$
and bias $\mathbf{b} \in \mathbb{R}^{C_{\mathrm{out}}}$.
Let
$\mathbf{m} \in \{0,1\}^{C_{\mathrm{out}} \times C_{\mathrm{in}} \times k \times k}$
be the binary pruning mask, so the surviving weight is
$\widetilde{\mathbf{W}} = \mathbf{m} \odot \mathbf{W}$.

For output channel $i$, the dense pre-activation on input
$\mathbf{x}$ is
\begin{equation}
    z_i = \langle \mathbf{w}_i,\, \mathbf{x} \rangle + b_i,
\label{eq:supp_dense_preact}
\end{equation}
where $\mathbf{w}_i$ denotes the $i$-th output filter flattened over
$C_{\mathrm{in}} \times k \times k$ dimensions. After pruning, the same
computation becomes
\begin{equation}
    \widetilde{z}_i
    =
    \langle \mathbf{m}_i \odot \mathbf{w}_i,\, \mathbf{x} \rangle + b_i
    =
    z_i -
    \langle (1-\mathbf{m}_i) \odot \mathbf{w}_i,\, \mathbf{x} \rangle .
\label{eq:supp_pruned_preact}
\end{equation}
Let
\[
    e_i =
    \langle (1-\mathbf{m}_i) \odot \mathbf{w}_i,\, \mathbf{x} \rangle
\]
denote the removed-weight contribution for channel $i$. The pruning
residual depends on both the removed weights and the input patch
distribution. Under a simplifying approximation in which input elements
are weakly correlated with the removed weights, its variance scales with
the removed weight energy,
\begin{equation}
    \mathrm{Var}(e_i)
    \approx
    \|(1-\mathbf{m}_i) \odot \mathbf{w}_i\|_2^2 \, \sigma_x^2 ,
\label{eq:supp_residual_var}
\end{equation}
where $\sigma_x^2$ denotes an average input variance. This expression is
not intended as an exact generative model of pruning damage. Rather, it
highlights that channels losing different amounts of weight energy can
experience different magnitudes of activation perturbation.

The pruned activation variance can be written as
\begin{equation}
    \mathrm{Var}(\widetilde{z}_i)
    =
    \mathrm{Var}(z_i - e_i)
    =
    \mathrm{Var}(z_i)
    +
    \mathrm{Var}(e_i)
    -
    2\,\mathrm{Cov}(z_i,e_i).
\label{eq:supp_var_identity}
\end{equation}
Thus pruning can either contract or distort activation variance
depending on the covariance term. In the high-sparsity regimes studied
in the main text, we empirically observe that many channels have strongly
reduced post-pruning variance, while other channels retain a substantial
fraction of their dense activation scale. This non-uniformity is the
key point: global magnitude pruning does not enforce uniform damage
across channels, so the resulting activation distribution shift can be
highly heterogeneous within a single layer.

\subsection{Channel-wise Repair with Data-driven Shrinkage}
\label{sec:supp_channel_repair}

Given dense and pruned pre-BatchNorm activations, a natural local repair
is to match the activation variance of each output channel. If a pruned
channel is rescaled by a scalar $\gamma_i$, its variance becomes
\[
    \mathrm{Var}(\gamma_i \widetilde{z}_i)
    =
    \gamma_i^2 \mathrm{Var}(\widetilde{z}_i).
\]
Matching this variance to the dense channel variance gives the population
target
\begin{equation}
    \gamma_i^\star
    =
    \sqrt{
    \frac{\mathrm{Var}(z_i)}
         {\mathrm{Var}(\widetilde{z}_i)}
    } .
\label{eq:supp_gamma_optimal}
\end{equation}
Equivalently, $\gamma_i^\star$ minimises
\[
    \left(
    \gamma_i^2 \mathrm{Var}(\widetilde{z}_i)
    -
    \mathrm{Var}(z_i)
    \right)^2
\]
over $\gamma_i \geq 0$ whenever the variances are positive.

Since the population variances are unknown, ASR estimates them from a
small calibration set. Let
$\mathbf{Y}_d \in \mathbb{R}^{N \times C \times H \times W}$ and
$\mathbf{Y}_p \in \mathbb{R}^{N \times C \times H \times W}$ denote the
pre-BatchNorm activations of a layer from the dense and pruned models,
computed on $N$ calibration images. The raw channel-wise
variance-matching estimator is
\begin{equation}
    \hat{\gamma}_i
    =
    \sqrt{
    \frac{\widehat{\mathrm{Var}}(\mathbf{Y}_{d,i})}
         {\widehat{\mathrm{Var}}(\mathbf{Y}_{p,i}) + \epsilon}
    },
\label{eq:supp_gamma_raw}
\end{equation}
where the variance is computed over batch and spatial dimensions
$(n,h,w)$, and $\epsilon>0$ is a numerical floor.

Raw channel-wise matching is unstable for nearly collapsed channels. If
$\widehat{\mathrm{Var}}(\mathbf{Y}_{p,i})$ is close to zero, then
$\hat{\gamma}_i$ can become large even when the residual activation is
mostly noise or numerical fluctuation. A layer-wise correction has the
opposite problem: it averages over all channels and applies the same
scalar to healthy and damaged channels alike. ASR addresses both issues
by estimating the correction at channel granularity and then shrinking
unreliable channel-wise estimates toward the identity.

We define a layer-level shrinkage baseline from the empirical
distribution of post-pruning channel variances,
\begin{equation}
    \lambda
    =
    \mathrm{median}
    \left(
    \left\{
    \widehat{\mathrm{Var}}(\mathbf{Y}_{p,i})
    \right\}_{i=1}^{C}
    \right).
\label{eq:supp_prior}
\end{equation}
The median is used as a robust layer-level scale. It is less affected
than the mean by a small number of anomalously large or small channel
variances. We then define
\begin{equation}
    s_i
    =
    \frac{
    \widehat{\mathrm{Var}}(\mathbf{Y}_{p,i})
    }{
    \widehat{\mathrm{Var}}(\mathbf{Y}_{p,i}) + \lambda
    },
\label{eq:supp_shrinkage}
\end{equation}
and set the final ASR correction factor to
\begin{equation}
    \gamma_i
    =
    s_i \hat{\gamma}_i + (1-s_i)\cdot 1 .
\label{eq:supp_gamma_final}
\end{equation}
Equivalently,
\begin{equation}
    \gamma_i - 1 = s_i(\hat{\gamma}_i - 1).
\label{eq:supp_gamma_interp}
\end{equation}
This form makes the reliability interpretation explicit. Channels with
small post-pruning variance have small $s_i$ and are pulled toward the
identity correction. Channels with sufficiently large post-pruning
variance have $s_i$ closer to one and retain more of the raw
variance-matching correction.

The terminology ``empirical Bayes'' is used in this limited sense:
the shrinkage scale is estimated from the empirical distribution of
post-pruning channel variances within the layer, and the resulting rule
contracts uncertain channel-wise estimates toward a conservative
identity target. We do not perform full Bayesian posterior inference
over weights or activations.

The weight update applies the correction per output channel:
\begin{equation}
    \widetilde{\mathbf{W}}_{\ell,i,:,:,:}
    \leftarrow
    \gamma_i \widetilde{\mathbf{W}}_{\ell,i,:,:,:},
    \qquad
    i=1,\ldots,C_{\mathrm{out}}.
\label{eq:supp_weight_update}
\end{equation}
This is equivalent to multiplying the corresponding pre-BatchNorm output
channel by $\gamma_i$.

\subsection{Formal Stability Analysis}
\label{sec:stability}

We now formalise the difference between a shared layer-wise correction
and the shrinkage-stabilised channel-wise correction used by ASR. The
purpose of this analysis is to capture the failure mode observed
empirically in the main text, not to claim a complete probabilistic
model of pruning.

\paragraph{Sensitivity of layer-wise scaling.}

Let a layer have $C$ output channels. Denote the post-pruning activation
variance of channel $i$ by
$v_i=\widehat{\mathrm{Var}}(\mathbf{Y}_{p,i})$, and the average dense
activation variance by
\[
    \bar{v}_d
    =
    \frac{1}{C}
    \sum_{i=1}^{C}
    \widehat{\mathrm{Var}}(\mathbf{Y}_{d,i}).
\]
Suppose the first $k$ channels are near-dead, so $v_i \leq \delta$ for
$i\leq k$, where $\delta>0$ is small. The layer-wise correction factor is
\begin{equation}
    \gamma_{\mathrm{LW}}^2
    =
    \frac{\bar{v}_d}
    {
    \frac{1}{C}
    \left(
    \sum_{i=1}^{k} v_i
    +
    \sum_{i=k+1}^{C} v_i
    \right)
    } .
\label{eq:supp_lw_ratio}
\end{equation}
As $\delta \to 0$, the contribution of the near-dead channels to the
denominator vanishes, and
\begin{equation}
    \gamma_{\mathrm{LW}}^2
    \to
    \frac{C\bar{v}_d}
    {
    \sum_{i=k+1}^{C} v_i
    } .
\label{eq:supp_lw_limit}
\end{equation}
Equation~\eqref{eq:supp_lw_limit} shows that the shared correction is
sensitive to the remaining active variance mass. If many channels are
collapsed, or if the remaining active channels have insufficient
variance to stabilise the layer average, the layer-wise scalar can grow
large. Since the same $\gamma_{\mathrm{LW}}$ is then applied to all
channels, the correction cannot distinguish healthy channels from
channels whose post-pruning signal is unreliable. This is the
granularity mismatch motivating ASR.

\paragraph{Shrinkage property of ASR.}

For ASR, the correction factor satisfies
\[
    \gamma_i = s_i \hat{\gamma}_i + (1-s_i),
    \qquad
    s_i=\frac{v_i}{v_i+\lambda}.
\]
Assume $\lambda>0$. If channel $i$ is near-dead with $v_i\leq\delta$,
then
\begin{equation}
    s_i
    =
    \frac{v_i}{v_i+\lambda}
    \leq
    \frac{\delta}{\delta+\lambda}
    \to 0
    \qquad
    \text{as } \delta\to 0.
\label{eq:supp_shrink_dead}
\end{equation}
Consequently,
\begin{equation}
    \gamma_i \to 1
    \qquad
    \text{as } v_i\to 0.
\label{eq:supp_identity_limit}
\end{equation}
Thus ASR does not apply a large raw variance-matching correction to
channels whose post-pruning activation variance has nearly vanished.

More generally, Eq.~\eqref{eq:supp_gamma_final} implies that each ASR
correction is a convex combination of the raw estimate and the identity:
\begin{equation}
    \gamma_i \in
    \left[
    \min\{1,\hat{\gamma}_i\},
    \max\{1,\hat{\gamma}_i\}
    \right].
\label{eq:supp_convex_bound}
\end{equation}
This property is the stability mechanism used by ASR. It does not make
the raw estimator itself bounded, nor does it guarantee that every
healthy channel is perfectly repaired. Instead, it ensures that the
applied correction is no more extreme than the raw estimate and that the
most unreliable low-variance channels are pulled toward no correction.

\paragraph{Degenerate zero-median case.}

The analysis above assumes $\lambda>0$. If the median post-pruning
variance in a layer is exactly zero, then at least half of the channels
have zero or numerically negligible variance on the calibration set. In
this degenerate case, the layer contains too little reliable
channel-wise information for aggressive multiplicative repair to be
well-conditioned. In implementation, this case should be handled
conservatively, for example by using a small numerical floor for
$\lambda$ or by falling back toward the identity correction for the
affected layer. This behaviour is consistent with the central design
principle of ASR: collapsed channels should not be amplified in an
attempt to match dense variance.

\subsection{Bias Correction and Integration with BatchNorm Recalibration}
\label{sec:supp_bias_bn}

Pruning can shift activation means as well as variances. After
channel-wise rescaling, we optionally adjust the convolutional bias so
that the repaired pre-BatchNorm activations match the dense activations
at the first moment. Let
$\boldsymbol{\gamma}=(\gamma_1,\ldots,\gamma_C)^\top$, and denote by
$\widehat{\mathbb{E}}[\mathbf{Y}_d]$ and
$\widehat{\mathbb{E}}[\mathbf{Y}_p]$ the per-channel sample means over
batch and spatial dimensions. If the repaired pre-BatchNorm activations
are written as
\[
    \widetilde{\mathbf{Y}}_p
    =
    \boldsymbol{\gamma}\odot \mathbf{Y}_p
    +
    (\mathbf{b}_{\mathrm{new}}-\mathbf{b}_{\mathrm{old}}),
\]
then imposing
\[
    \widehat{\mathbb{E}}[\widetilde{\mathbf{Y}}_p]
    =
    \widehat{\mathbb{E}}[\mathbf{Y}_d]
\]
gives
\begin{equation}
    \mathbf{b}_{\mathrm{new}}
    =
    \mathbf{b}_{\mathrm{old}}
    +
    \widehat{\mathbb{E}}[\mathbf{Y}_d]
    -
    \boldsymbol{\gamma}
    \odot
    \widehat{\mathbb{E}}[\mathbf{Y}_p].
\label{eq:supp_bias}
\end{equation}

For networks with BatchNorm layers, ASR is applied before BatchNorm
recalibration. ASR estimates channel-wise repair factors from 64
calibration images using forward evaluation only. BatchNorm
recalibration is then performed under the same recalibration budget used
for all repair baselines in the corresponding experiment. Thus ASR does
not use gradient computation or sparse-model retraining, and it does not
change the pruning mask.

\section{Choice of Calibration Step for Main Reporting}
\label{sec:step20}

We use calibration step $b{=}20$ as the primary reporting point for
BN Only, LW+BN, and ASR+BN. This choice reflects a practical trade-off.
By $b{=}20$, BatchNorm recalibration has already recovered a substantial
portion of its eventual performance across datasets, while the gap
between methods remains visible enough to distinguish the effect of
post-pruning repair. At smaller calibration budgets, BN statistics are
often still too noisy for a stable comparison. At larger budgets, the
marginal gains from additional recalibration become smaller and the
performance of different repair methods begins to compress toward a
common ceiling.

\begin{figure}[t]
    \centering
    \includegraphics[width=\linewidth]{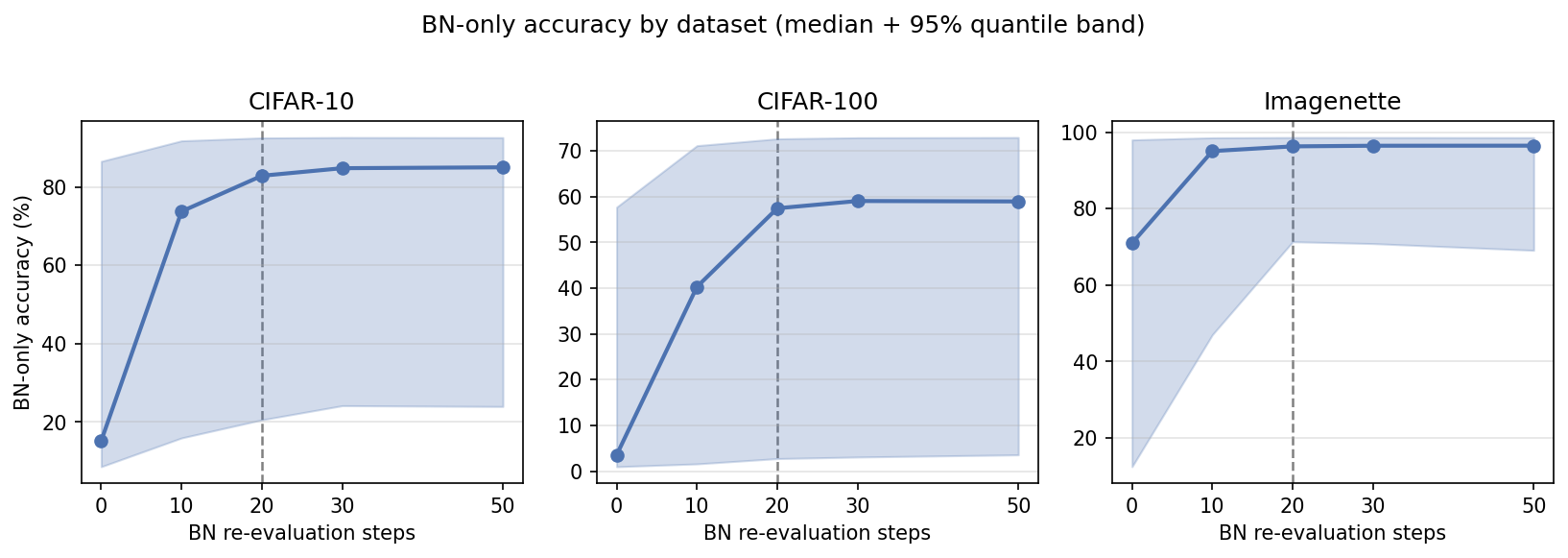}
    \caption{BN Only top-1 accuracy (median and 95\% quantile band)
    as a function of BatchNorm recalibration steps, shown separately
    for each dataset. The dashed vertical line marks step $b{=}20$,
    which captures most of the recoverable gain while retaining useful
    separation among repair methods.}
    \label{fig:step20_rationale}
\end{figure}

Table~\ref{tab:bn-re-eval-acc-vs-step} and
Figure~\ref{fig:step20_rationale} together illustrate this trade-off.
Across datasets, step $20$ captures most of the recoverable benefit of
BN recalibration while remaining early enough to preserve sensitivity to
differences among repair methods. For this reason, it provides a stable
and informative operating point for the main comparison.

\begin{table}[t]
    \centering
    \scriptsize
    \setlength{\tabcolsep}{3pt}
    \renewcommand{\arraystretch}{1.15}
    \caption{Top-1 accuracy (\%) versus BatchNorm recalibration step.
    Each entry is reported as median [min, max] across evaluated
    settings. Step $20$ is used as the primary reporting point in the
    main text.}
    \label{tab:bn-re-eval-acc-vs-step}
    \begin{tabular}{lccccc}
    \toprule
    Dataset & Step 0 & Step 10 & Step 20 & Step 30 & Step 50 \\
    \midrule
    CIFAR-10
    & \shortstack{15.20 \\ {\footnotesize [8.60, 86.54]}}
    & \shortstack{73.75 \\ {\footnotesize [15.94, 91.78]}}
    & \shortstack{\textbf{82.86} \\ {\footnotesize [20.57, 92.53]}}
    & \shortstack{84.79 \\ {\footnotesize [24.17, 92.63]}}
    & \shortstack{85.02 \\ {\footnotesize [23.97, 92.60]}} \\[2pt]
    CIFAR-100
    & \shortstack{3.52 \\ {\footnotesize [0.96, 57.72]}}
    & \shortstack{40.30 \\ {\footnotesize [1.58, 71.17]}}
    & \shortstack{\textbf{57.47} \\ {\footnotesize [2.73, 72.67]}}
    & \shortstack{59.06 \\ {\footnotesize [3.07, 72.91]}}
    & \shortstack{58.97 \\ {\footnotesize [3.55, 72.96]}} \\[2pt]
    Imagenette
    & \shortstack{70.95 \\ {\footnotesize [12.50, 97.98]}}
    & \shortstack{95.04 \\ {\footnotesize [47.03, 98.54]}}
    & \shortstack{\textbf{96.28} \\ {\footnotesize [71.34, 98.61]}}
    & \shortstack{96.45 \\ {\footnotesize [70.85, 98.60]}}
    & \shortstack{96.46 \\ {\footnotesize [69.10, 98.57]}} \\
    \midrule
    Overall
    & \shortstack{23.16 \\ {\footnotesize [1.00, 97.77]}}
    & \shortstack{66.01 \\ {\footnotesize [2.04, 98.26]}}
    & \shortstack{\textbf{79.20} \\ {\footnotesize [5.89, 98.33]}}
    & \shortstack{80.76 \\ {\footnotesize [6.46, 98.36]}}
    & \shortstack{80.74 \\ {\footnotesize [7.04, 98.37]}} \\
    \bottomrule
    \end{tabular}
\end{table}

\section{Robustness Checks and Additional Plots}
\label{sec:additional}

\subsection{Calibration-step trends on CIFAR-100}
\label{sec:supp_cifar100_trends}

\begin{figure}[t]
    \centering
    \includegraphics[width=\linewidth]{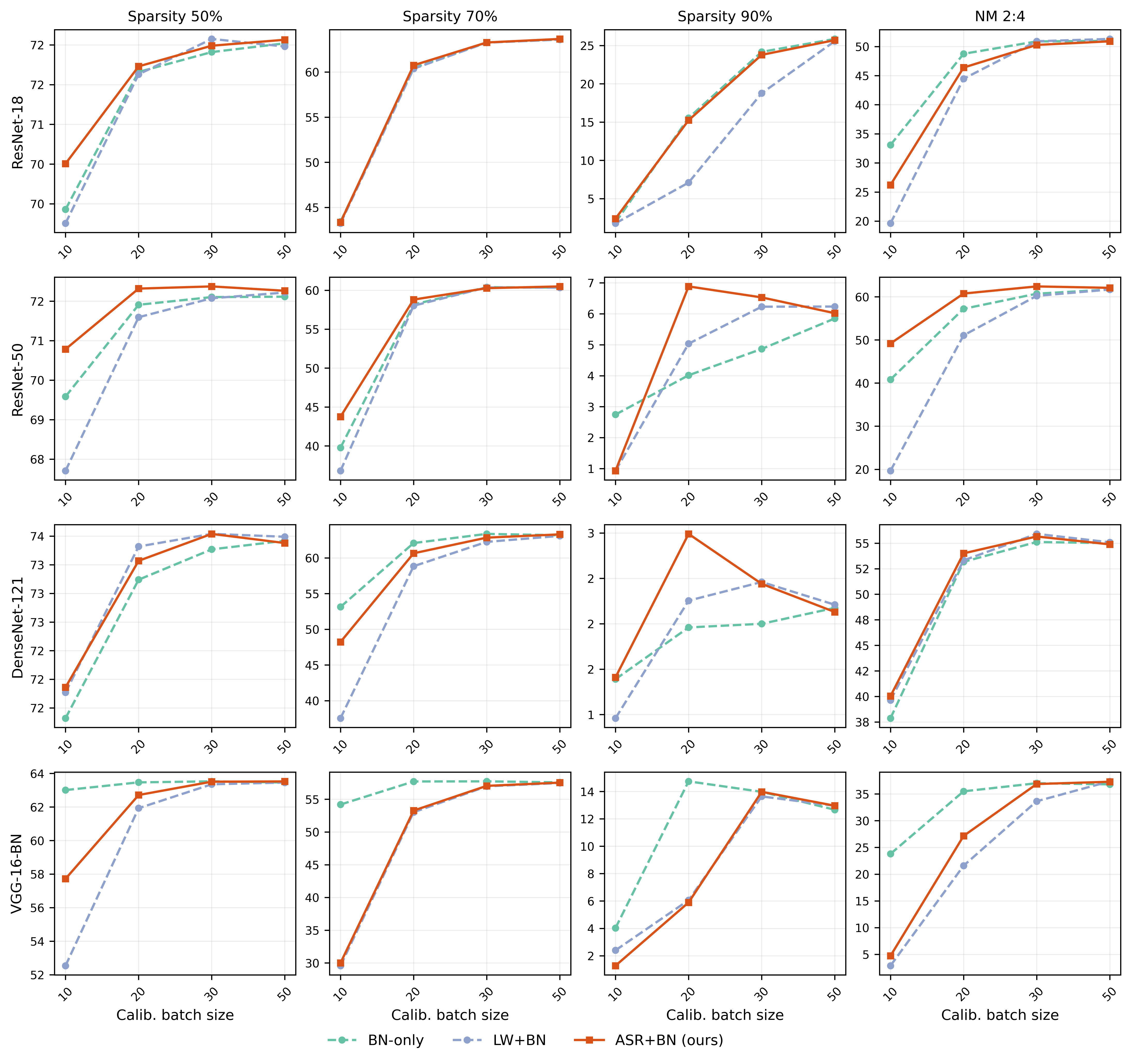}
    \caption{Top-1 accuracy versus calibration batch size on CIFAR-100.
    The overall trend is consistent with CIFAR-10, with ASR+BN showing
    the clearest advantage over LW+BN in the higher-sparsity settings.}
    \label{fig:cifar100}
\end{figure}

Figure~\ref{fig:cifar100} complements the CIFAR-10 calibration curves in
Figure~\ref{fig:cifar10} of the main paper by showing top-1 accuracy
versus calibration batch size on CIFAR-100. The pattern is broadly consistent with CIFAR-10:
ASR+BN shows its clearest advantage over LW+BN in the higher-sparsity
settings, especially at 90\% sparsity and under NM~2:4 sparsity, while
differences among methods narrow as the calibration budget increases.

\subsection{Pruning Severity and Repair Gap}

\begin{figure}[t]
    \centering
    \includegraphics[width=\linewidth]{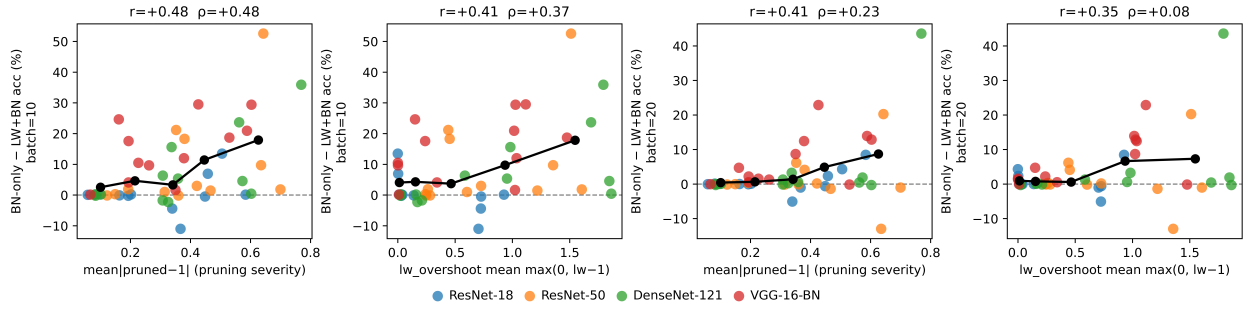}
    \caption{Accuracy gap between BN Only and LW+BN as a function of
    pruning severity (left column: mean$|\gamma_{\mathrm{pruned}}-1|$;
    right column: layer-wise overshoot mean$\max(0,\gamma_{\mathrm{LW}}-1)$)
    at calibration steps $b{=}10$ (top) and $b{=}20$ (bottom).
    Larger severity is associated with a larger advantage for BN Only
    over layer-wise repair, consistent with the instability identified
    in Section~\ref{sec:stability}.}
    \label{fig:scatter_bn_lw}
\end{figure}

Figure~\ref{fig:scatter_bn_lw} extend the
correlation analysis in Section~5.3 of the main text by reporting results
at both $b{=}10$ and $b{=}20$, and by including a second severity
measure: the layer-wise overshoot $\mathrm{mean}\max(0, \gamma_{\mathrm{LW}}-1)$,
which directly quantifies how much layer-wise scaling exceeds the
identity on average.

Figure~\ref{fig:scatter_bn_lw} shows the accuracy gap between BN Only
and LW+BN as a function of both severity measures. The positive
correlation confirms that layer-wise repair degrades relative to
BN Only most strongly when pruning induces large variance distortion or
large overshoot, consistent with the instability analysis in
Section~\ref{sec:stability}.

Figure~\ref{fig:scatter_bn_lw} shows the corresponding gap between
ASR+BN and LW+BN. At $b{=}20$ (bottom-left panel), the correlation with
pruning severity reaches $r{=}0.61$, $\rho{=}0.70$, matching the value
reported in the main text. At $b{=}10$ the correlation is weaker
($r{=}0.19$, $\rho{=}0.33$), reflecting the additional noise in BN
statistics at very small calibration budgets.

\subsection{Per-layer Variance Statistics}
\label{sec:supp_layer_stats}

\begin{figure}[t]
    \centering
    \includegraphics[width=\linewidth]{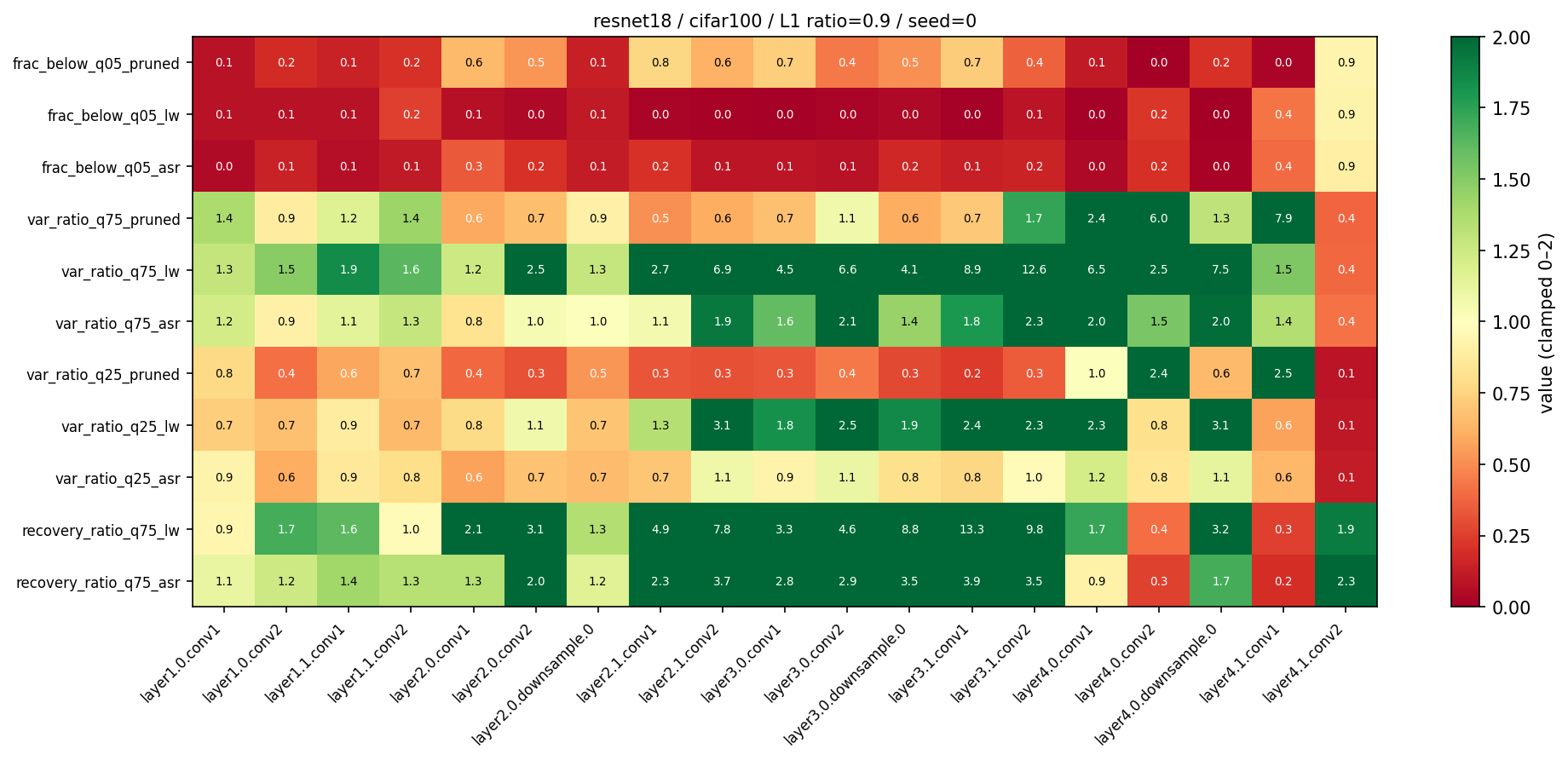}
    \caption{Per-layer channel variance statistics for ResNet-18 on
    CIFAR-100 at 90\% global L1 sparsity. Values clamped to $[0,2]$.
    Layer-wise repair over-corrects in the deeper layers
    (\texttt{var\_ratio\_q75\_lw} up to 12.6); ASR remains closer
    to the dense reference throughout.}
    \label{fig:heatmap_resnet18}
\end{figure}

\begin{figure}[t]
    \centering
    \includegraphics[width=\linewidth]{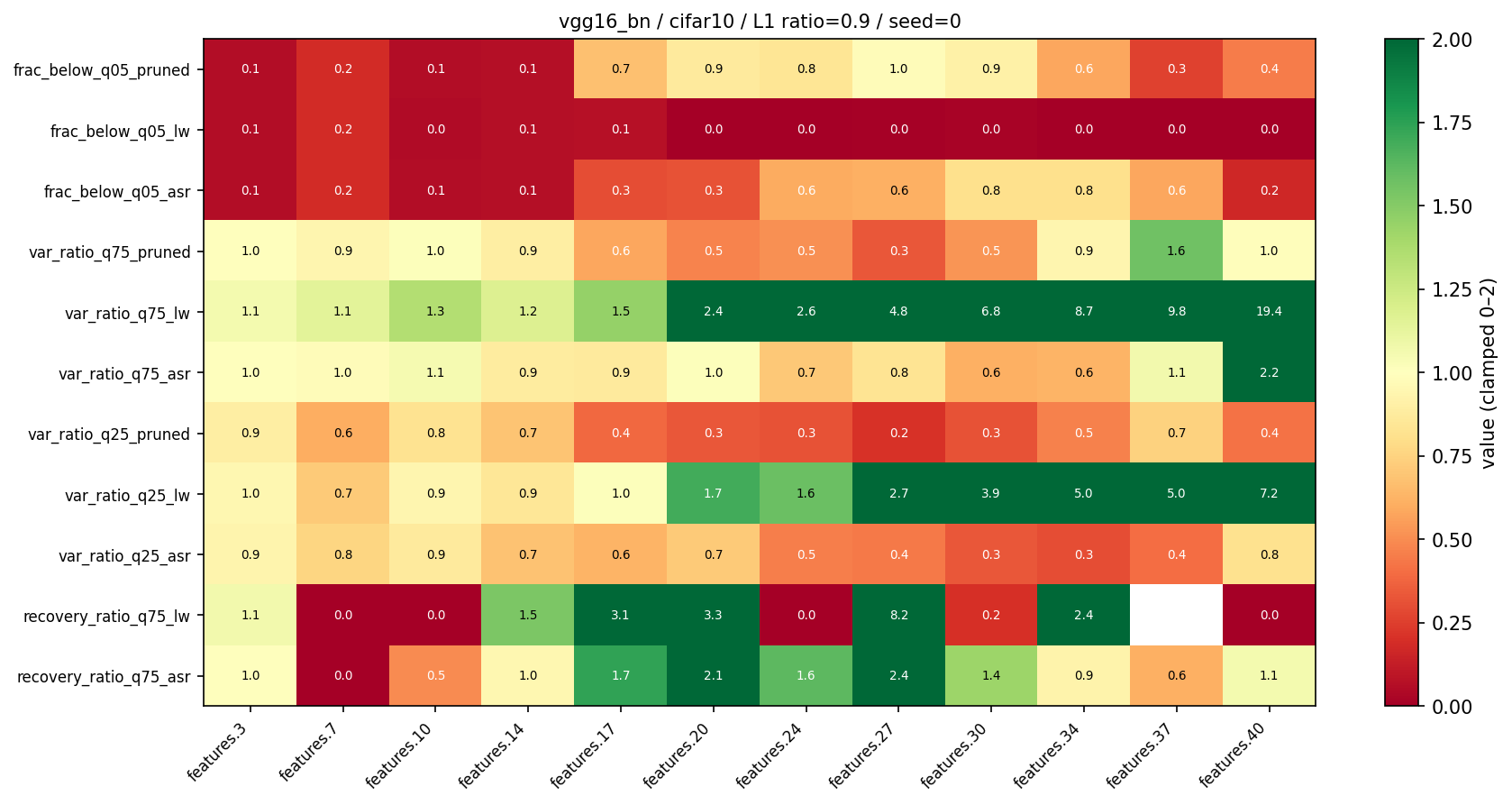}
    \caption{Per-layer channel variance statistics for VGG-16-BN on
    CIFAR-10 at 90\% global L1 sparsity. Over-correction is more
    severe than in ResNet-18, with \texttt{var\_ratio\_q75\_lw}
    reaching 19.4 at \texttt{features.40}, consistent with
    BN Only outperforming both repair methods on this architecture.}
    \label{fig:heatmap_vgg16}
\end{figure}

Figures~\ref{fig:heatmap_resnet18} and~\ref{fig:heatmap_vgg16} show
per-layer heatmaps of channel variance statistics for ResNet-18 on
CIFAR-100 and VGG-16-BN on CIFAR-10, both at 90\% global L1 sparsity.
Each row corresponds to one diagnostic statistic (fraction of channels
below the 5th percentile of dense variance, 25th and 75th percentile
variance ratios, and recovery ratios for LW+BN and ASR+BN), and each
column corresponds to one convolutional layer.

For ResNet-18 (Figure~\ref{fig:heatmap_resnet18}), the
\texttt{var\_ratio\_q75\_lw} row shows values well above 1 in the
deeper layers (up to 12.6 at \texttt{layer3.1.conv1}), indicating
systematic over-correction by layer-wise repair. The corresponding
\texttt{var\_ratio\_q75\_asr} row remains much closer to 1 throughout,
confirming that ASR suppresses the overshoot. For VGG-16-BN
(Figure~\ref{fig:heatmap_vgg16}), over-correction is even more
pronounced in the later feature layers (up to 19.4 at
\texttt{features.40}), which is consistent with the observation in the
main text that BN Only outperforms both repair methods on this
architecture at 90\% sparsity.

\subsection{Additional Calibration Curves}
\label{sec:supp_nm24_curves}

\begin{figure}[t]
    \centering
    \includegraphics[width=\linewidth]{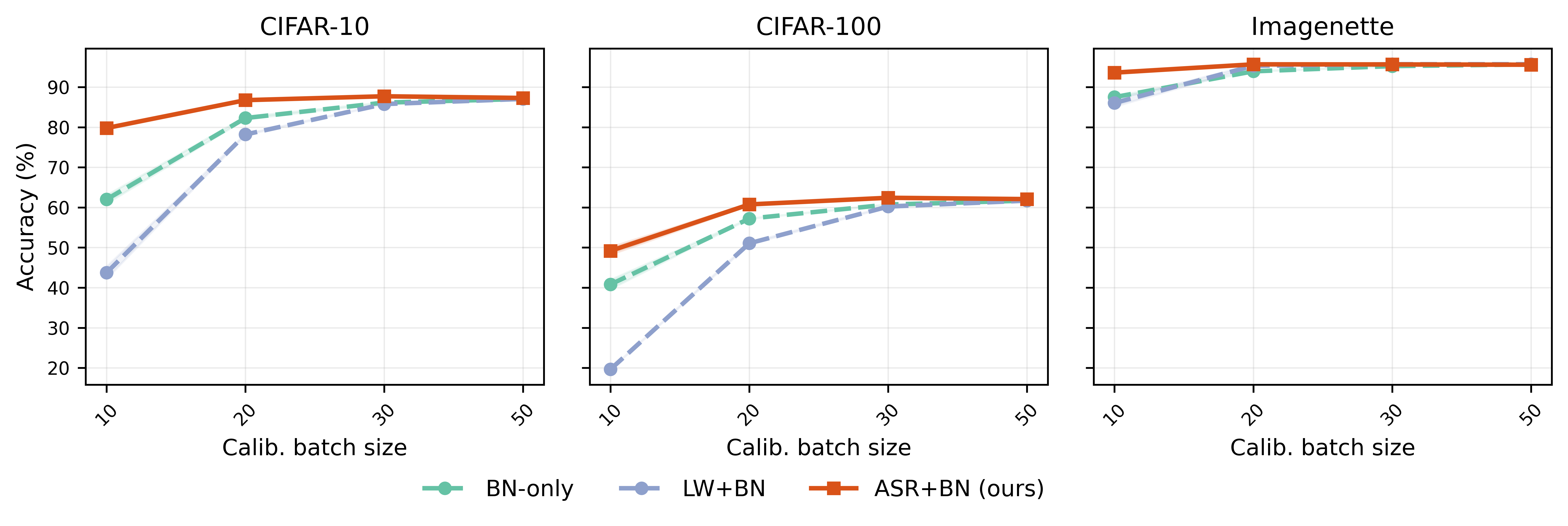}
    \caption{Top-1 accuracy versus calibration batch size for ResNet-50
    under NM~2:4 structured sparsity on CIFAR-10, CIFAR-100, and
    Imagenette. ASR+BN shows the largest advantage at small calibration
    budgets on the two harder datasets.}
    \label{fig:resnet50_nm24}
\end{figure}

Figure~\ref{fig:resnet50_nm24} extends the structured-sparsity comparison
in Table~\ref{tab:main} by showing top-1 accuracy versus calibration
batch size for ResNet-50 under NM~2:4 structured sparsity across all
three datasets. ASR+BN leads BN Only and LW+BN at small calibration
budgets on CIFAR-10 and CIFAR-100, with the advantage most visible at
$b{=}10$. On Imagenette, where the fine-tuned dense model is close to
ceiling performance, all three methods converge rapidly and the
differences are small, consistent with the pattern observed in the main
text.

\end{document}